%% file: conference.tex
\documentclass{article} 
\usepackage{conference,times}

\input{math_commands.tex}

\usepackage{hyperref}
\usepackage{url}
\usepackage{xspace}
\usepackage{graphicx}
\usepackage{wrapfig}
\usepackage{booktabs}
\usepackage{multirow}
\usepackage{caption}
\usepackage{wrapfig}
\usepackage{bbm}
\usepackage{subcaption}

\title{Less Context, Better Geometry: Masked Geometric Encoder for Robust 3D Foundation Models}
\iclrfinalcopy

\author{Zhimin Shao, Xijun Liu, Zhaoliang Zhang, Yutao Tang, Abhay Yadav, Rama Chellappa \\
Department of Electrical and Computer Engineering\\
Johns Hopkins University\\
\texttt{\{zshao14, xliu253, zzhan88, ytang67, ayadav13, rchella4\}@jhu.edu} \\
\And
Cheng Peng \\
Department of Data Science \\
University of Virginia \\
\texttt{cheng.peng@virginia.edu} \\
}

\begin{document}

\maketitle
\fancyhead[L]{Preprint}
\begin{abstract}
Recent progress in 3D foundation models has enabled rapid 3D reconstruction and camera calibration by leveraging learned 3D priors from vast amount of spatial data. However, the all-to-all global attention design leads to quadratic complexity and limits long-sequence inference; unconstrained cross-view interactions also can propagate unreliable evidence from occluded or visually similar but geometrically distant views.
In this paper, 
We introduce a \emph{Masked Geometric Encoder} (MGE), which promotes the learning of robust geometric representations under incomplete cross-view context. 
During training, MGE strategically drops frame tokens from global attention and distills from a pretrained full-context teacher model. This allows the model to learn an intrinsically richer per-frame representation while providing sufficient intermediate supervision to avoid performance degradation.
Through extensive experiments, we show that MGE leads to much stronger performance under occlusion and doppelganger views while retaining high performance on standard benchmarks. Such a richer frame representation also leads to more effective token reduction during inference. To this end, we develop a novel \emph{Anchor-Guided Adaptive} token merging technique that preserves representative anchor frames while jointly merging redundant tokens from the remaining views. Compared to other efficient inference approaches, we can achieve inference speedup while consistently maintaining higher reconstruction quality, particularly in limited-view settings.

\end{abstract}

\input{sec/1_intro}

\input{sec/2_related}
\input{sec/3_methods}
\input{sec/4_experiments}

\input{sec/5_conclusion}


\section*{Acknowledgement}
This research is based upon work supported by the Office of the Director of National Intelligence
(ODNI), Intelligence Advanced Research Projects Activity (IARPA), via IARPA R\&D Contract
No. 140D0423C0076. The views and conclusions contained herein are those of the authors and
should not be interpreted as necessarily representing the official policies or endorsements, either
expressed or implied, of the ODNI, IARPA, or the U.S. Government. The U.S. Government is
authorized to reproduce and distribute reprints for Governmental purposes notwithstanding any
copyright annotation thereon.

\subsection*{AI use statement}
In this work, we used generative AI tools to improve
the language and readability of this manuscript. All scientific contributions, including the core ideas, methodology, experimental design, and interpretation of results, are the original work of the authors. All authors have thoroughly reviewed the final version and
take full responsibility for its content and claims.

\bibliography{conference}
\bibliographystyle{conference}
\appendix
\input{sec/appendix}


\end{document}

%% file: math_commands.tex
\usepackage{amsmath,amsfonts,bm}

\def\eqref#1{equation~\ref{#1}}

\def\1{\bm{1}}

\DeclareMathAlphabet{\mathsfit}{\encodingdefault}{\sfdefault}{m}{sl}
\SetMathAlphabet{\mathsfit}{bold}{\encodingdefault}{\sfdefault}{bx}{n}



%% file: sec/1_intro.tex
\section{Introduction}

Extracting 3D concepts from images is a fundamental research problem in computer vision, with various applications in autonomous driving, geo-localization, augmented and virtual reality, and robotics~\citep{geiger2013vision, arandjelovic2016netvlad, newcombe2011kinectfusion, mur2015orb}. Recent feedforward 3D foundation models~\citep{Wang_2024_CVPR, wang2025vggt, wang2026vggt} have substantially advanced this capability. These models jointly estimate camera parameters and dense scene geometry from multiple images in one forward pass. By encoding geometric priors learned from large-scale 3D data, these models produce scene-level predictions without requiring the iterative, scene-specific optimization conventionally used in Structure-from-Motion (SfM) pipelines.

Despite impressive progress, scalability and robustness remain important challenges for current 3D foundation models. Conventional encoder design aggregates multi-view information through dense global attention. However, additional observations do not always provide reliable geometric evidence. For example, visually similar but geometrically distinct structures, often referred to as Doppelgangers~\citep{cai2023doppelgangers}, can induce incorrect associations across views. Furthermore, the computation grows quadratically with the total number of frames, making long sequences increasingly expensive to process.

Conventional SfM pipelines provide explicit mechanisms for dealing with these challenges. COLMAP~\citep{schoenberger2016sfm} constructs a view graph from selected image pairs and geometrically verified correspondences. As such, we can remove redundant or ambiguous matching pairs in the view graph through additional prior~\citep{schoenberger2016sfm, arandjelovic2016netvlad, berton2025megaloc, xiangli2025doppelgangers++}, leading to speedup or improved accuracy in calibration. Current 3D foundation models compute all-to-all attention across all frames. Naively removing or sparsifying these global interactions can lead to substantial performance degradation~\citep{wang2026block, sun2026avggt}, suggesting that the pretrained representations are highly dependent on dense cross-view connectivity. By drawing an analogy between a calibration view graph and global attention, we know that not every frame-to-frame relationship is required or can even hurt in bundle adjustment. This prompts our question: \textbf{Can 3D foundation models learn a more robust representation to missing frames?}

To this end, we introduce Masked Geometric Encoder (MGE), a flexible framework for learning robust geometric representations in feed-forward 3D foundation models. MGE draws inspiration from masked representation learning~\citep{he2022masked} and teacher--student learning~\citep{caron2021emerging,oquab2023dinov2}. Instead of masking image content, we randomly remove frame-to-frame connections during global attention, restricting the available cross-view information to the geometric encoder to simulate sparse calibration view graph. Importantly, the underlying scene geometry remains unchanged: the network must still recover consistent 3D structure from the remaining cross-view interactions. Training under varying connectivity patterns therefore discourages excessive reliance on particular view pairs and encourages each frame representation to retain more geometrically informative context.

To supervise learning under incomplete context, we employ a self-distillation strategy that maintains geometric knowledge from the pretrained base model to MGE. Instead of relying on direct ground-truth supervision, which can be sparse (as shown in the Appendix~\ref{app:sparse}), we align the masked student with the teacher at two complementary levels: intermediate geometric features and predicted geometry. The feature-level objective encourages the student to preserve the teacher’s geometric representation despite reduced cross-view access, while the output-level objective constrains its resulting 3D predictions including point maps and camera poses. Consequently, MGE learns to recover the same underlying geometry from different subsets of cross-view evidence, promoting representations that remain stable when individual view relations are missing or unreliable.

Finally, we connect this robust representation learning objective to efficient long-sequence inference through an Anchor-Guided Adaptive (AGA) token-merging strategy. Token merging~\citep{bolya2022token} reduces inference cost by combining similar or redundant tokens before global attention. Since MGE learns richer and more independent per-frame representations, it enables more effective merging than standard token-merging strategies~\citep{shen2025fastvggt}. During inference, AGA uniformly selects a sparse set of full-resolution anchor frames and use their features to estimate the redundancy of the remaining views. Non-anchor frames are then assigned adaptive merging ratios according to their similarity to the anchor context, such that highly redundant views are compressed more aggressively while complementary views retain more tokens. Compared to standard token merging, our approach leads to significant performance improvement. The advantage is especially pronounced when neighboring views are farther apart, where uniform token reduction~\citep{shen2025fastvggt} can discard geometric information and cause severe reconstruction degradation, whereas our method better preserves full-token performance. 


To summarize, our contributions are as follows:
\begin{enumerate}
    \item We introduce MGE, which uses random frame-feature dropping within global attention to train feedforward 3D multi-view encoders under incomplete cross-view context.
    \item During MGE training, we apply a self-distillation design from a pretrained base model through intermediate-feature and geometry alignment, providing supervision beyond sparse geometric ground truth that degrades performance.
    \item Building on the robust representations learned by MGE, we propose AGA token merging, achieving better reconstruction quality than existing token-merging methods at a comparable computational cost, particularly in limited-view regimes.
    \item We perform large scale experiments and demonstrate that MGE is significantly more robust and accurate in challenging evaluation settings such as occlusions, and doppelgangers compared to previous SoTA, while retaining top performance in conventional evaluations. 
\end{enumerate}

%% file: sec/2_related.tex
\section{Related Works}

\paragraph{Camera Calibration with View Graphs.}
Traditional structure-from-motion pipelines such as COLMAP~\citep{schoenberger2016sfm} estimate camera poses through pairwise feature matching~\citep{sun2021loftr, edstedt2024roma,edstedt2025roma,shao2026spider}, geometric verification, and bundle adjustment. These pipelines naturally form a \emph{view graph}, where edges encode reliable geometric relations between image pairs. The quality of this graph is critical: incorrect connections caused by visually similar but geometrically unrelated views can lead to erroneous camera estimates and corrupted reconstructions. A line of work therefore focuses on identifying and removing unreliable pairwise relations before geometric optimization. Doppelgangers~\citep{cai2023doppelgangers} and Doppelgangers++~\citep{xiangli2025doppelgangers++}, for example, detect visually ambiguous image pairs and prune spurious view-graph edges. These studies show that dense connectivity is not necessarily desirable: selectively restricting cross-view interactions can improve geometric consistency.

\paragraph{Feed-forward 3D Reconstruction.}
Recent methods replace explicit correspondence construction and geometric optimization with learned geometric prediction. DUSt3R~\citep{Wang_2024_CVPR} regresses dense pointmaps from uncalibrated image pairs, while MASt3R~\citep{leroy2024grounding} augments this formulation with learned descriptors and matching supervision. Moving beyond pairwise prediction, VGGT~\citep{wang2025vggt} jointly estimates cameras, depth, pointmaps, and tracks from multiple images, whereas CUT3R~\citep{wang2025continuous} maintains a recurrent scene representation for continuous geometric perception. In these models, pairwise geometric relations are no longer explicitly represented as view-graph edges, but are instead implicitly encoded through learned global attention. This shifts the robustness problem from selecting geometric constraints before optimization to learning how information should propagate across views.

Subsequent feed-forward approaches have also explored different organizations of multi-view inference. GLUEMAP~\citep{pan2026gluemap} combines local feed-forward reconstruction with global SfM, VGGT-Long~\citep{deng2025vggt} processes views in overlapping subsets, and Diversity-aware VGGT~\citep{park2026diversity} partitions views according to visual and spatial diversity. These approaches modify which views interact or how their predictions are combined at inference time. In contrast, MGE targets the learned interaction itself, training the geometric encoder under restricted cross-view connectivity so that its representation remains robust to missing or unreliable view relations.


\paragraph{Long-Sequence and Efficient 3D Foundation Models.}
Existing approaches scale reconstruction through joint prediction, incremental memory, and compressed attention. Fast3R predicts globally aligned geometry from large image collections, while Spann3R and MUSt3R accumulate cross-view information through spatial or multi-layer memory~\citep{yang2025fast3r,wang20253d, cabon2025must3r}. FlashVGGT and VGGT-$\Omega$ reduce cross-view computation through compact descriptor or scene tokens~\citep{wang2026flashvggt, wang2026vggt}. Other methods modify inference without retraining: STAC compresses the key–value cache for streaming reconstruction, while HTTM and FastVGGT reduce computation through token merging~\citep{wang2026stac, wang2026httm, shen2025fastvggt}. These methods improve scalability through architectural compression, memory management, or inference-time token reduction, with some also improving reconstruction quality. MGE complements these directions by explicitly learning geometric representations that remain reliable under partial cross-view connectivity, and AGA subsequently exploits this property to compress redundant frames adaptively during inference.



%% file: sec/3_methods.tex
\section{Methods}
\subsection{Preliminary: Multi-View Encoder in Feed-Forward 3D Reconstruction}
Recent feed-forward 3D foundation models~\citep{wang2025vggt, lin2025depth, wang2026vggt, keetha2026mapanything, wang2026pi} follow an encoder--decoder formulation. Given a set of $N$ input images $\mathcal{I}=\{I_i\}_{i=1}^{N}$, an image encoder, such as DINOv2, first maps each image into a sequence of visual tokens $\mathbf{X}_i \in \mathbb{R}^{M\times d}$, where $M$ is the number of tokens per image and $d$ is the feature dimension. 
A multi-view encoder then alternates between frame attention and global attention for $L$ layers. Frame attention updates tokens independently within each image, while global attention jointly processes tokens from all views:
\begin{equation}
    \mathbf{F}_i^l = \operatorname{SelfAttn}_{\mathrm{frame}}(\mathbf{G}_i^l),
    \quad
    \mathbf{G}^l = \operatorname{SelfAttn}_{\mathrm{global}}
    \left(
    [\mathbf{F}_1^l;\ldots;\mathbf{F}_N^l]
    \right), \quad l=1,\cdots, L,
    \label{eq:frame_global}
\end{equation}
where $\mathbf{G}_i^1=\mathbf{X}_i$. The output $\mathbf{G}^l$ is partitioned into per-frame representations $\{\mathbf{G}_i^l\}_{i=1}^{N}$,and used as the input to the next frame attention layer. 

By alternating frame and global attention, the multi-view encoder progressively integrates local appearance information with multi-view geometric context. The resulting per-frame representations $\mathbf{G}_i^L$ are then decoded by task-specific prediction heads to recover scene geometry and camera information, including point maps, depth maps, confidence maps, and camera poses.

\begin{figure}
    \centering
    \includegraphics[width=1\linewidth]{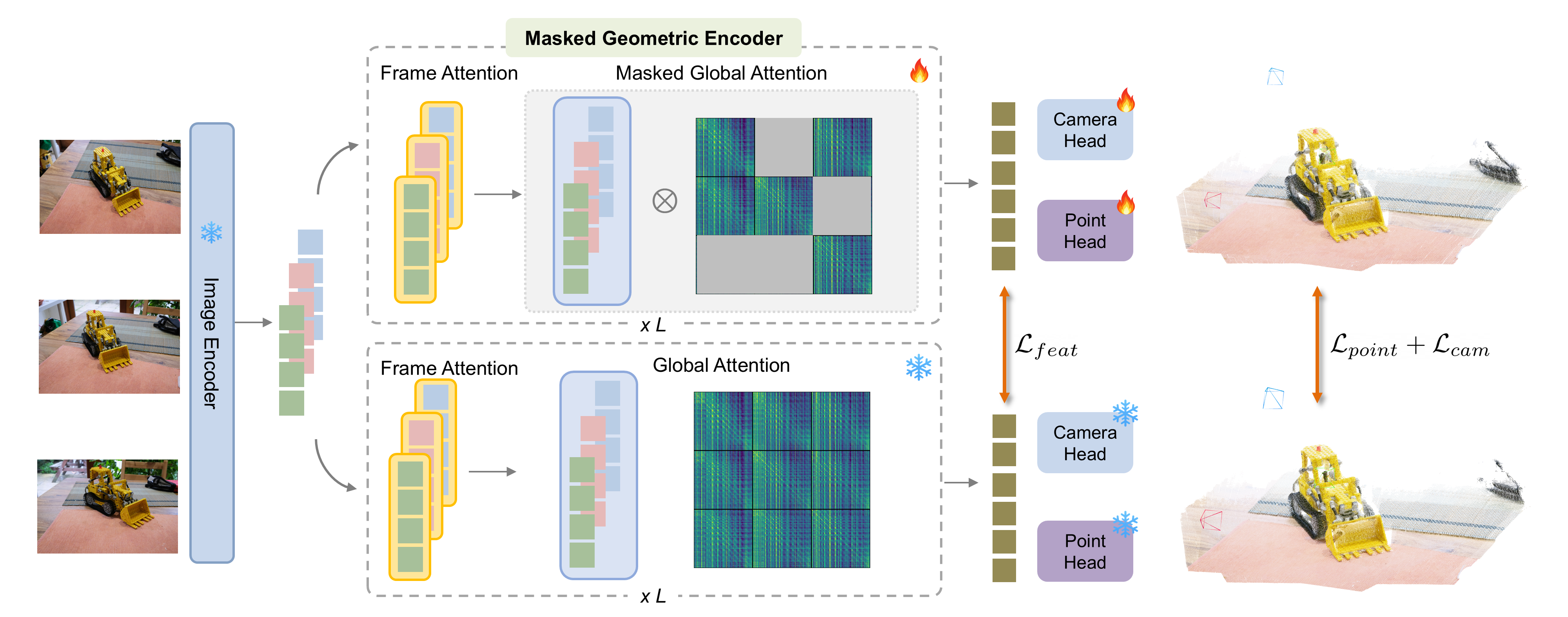}
    \caption{\textbf{Method overview.} During training, MGE masks the global attention randomly to limit the all-to-all cross-view information flow. To promote richer intermediate feature representation, we further introduce constrain MGE to produce similar 3D features compared to the full-attention teacher encoder output.}
    \label{fig:architecture}
    \vspace{-10pt}
\end{figure}
\subsection{Masked Geometric Encoder}
Global attention provides unrestricted cross-view information flow in conventional 3D foundation models; however, prior works~\citep{arandjelovic2016netvlad,cai2023doppelgangers, xiangli2025doppelgangers++} in camera calibration have demonstrated that eliminating certain cross-view pairs from bundle adjustment can be beneficial. In dense acquisition, many cross-view pairs are redundant, while some pairs can be harmful due to ambiguous features. By removing these pairs from a view graph, we can obtain better efficiency and higher performance. 

As shown in Fig~\ref{fig:architecture}, to improve robustness to unreliable cross-view information and and overall efficiency, we introduce a \emph{Masked Geometric Encoder} that explicitly regularizes frame-to-frame attention during training. 
Let $\mathbf{A}\in\{0,1\}^{N\times N}$ denote a frame-level connectivity mask, where $\mathbf{A}_{ij}=1$ indicates that view $i$ is allowed to attend to view $j$, and $\mathbf{A}_{ij}=0$ indicates that the corresponding cross-view interaction is removed. 
During training, we randomly generate $\mathbf{A}$ according to a predefined mask ratio $\rho$. Specifically, for each off-diagonal entry, we sample
\begin{equation}
    \mathbf{A}_{ij} \sim \operatorname{Bernoulli}(1-\rho),
    \qquad i\neq j.
\end{equation}
The diagonal entries are always preserved, i.e., $\mathbf{A}_{ii}=1$, 
so that each frame can attend to itself. We note that $\mathbf{A}$ can be realized in forms other than a Bernoulli distribution, and provide additional experiments on other realizations in the Appendix~\ref{app:ablate_pattern}.

The frame-level mask is expanded to the token level and applied to the global-attention operation. Given query, key, and value matrices $\mathbf{Q}$, $\mathbf{K}$, and $\mathbf{V}$, the masked global attention is defined as
\begin{equation}
    \operatorname{MaskAttn}_{\mathbf{A}}
    (\mathbf{Q},\mathbf{K},\mathbf{V})
    =
    \operatorname{Softmax}
    \left(
    \frac{\mathbf{Q}\mathbf{K}^{\top}}{\sqrt{d}}
    +
    \mathbf{M}_{\mathbf{A}}
    \right)
    \mathbf{V},
\end{equation}
where $\mathbf{M}_{\mathbf{A}}$ is the token-level attention mask induced by the frame-level connectivity matrix $\mathbf{A}$. For a query token belonging to frame $i$ and a key token belonging to frame $j$, the corresponding mask value is
\begin{equation}
    (\mathbf{M}_{\mathbf{A}})_{ij}
    =
    \begin{cases}
        0, & \text{if } \mathbf{A}_{ij}=1,\\
        -\infty, & \text{if } \mathbf{A}_{ij}=0.
    \end{cases}
\end{equation}
The masking operation is applied exclusively to global attention, while frame attention remains unchanged. Specifically, we replace the standard self-attention operation for global attention in Eq.~\ref{eq:frame_global} with the masked attention operator defined above:
\begin{equation}
    \mathbf{G}^{(l)}
    =
    \operatorname{MaskAttn}_{\mathbf{A}}
    \left(
    [\mathbf{F}_1^{(l)};\ldots;\mathbf{F}_N^{(l)}]
    \right).
\end{equation}
Therefore, each image can still fully model its intra-frame spatial structure, while cross-frame information exchange is restricted according to the sampled connectivity pattern. During training, different frame-to-frame connections are removed across iterations, forcing the encoder to reconstruct consistent 3D geometry from varying subsets of cross-view information. 
MGE is a model-agnostic framework that can be integrated into feed-forward 3D architectures with cross-view attention. We use $\pi^3$ as the primary instantiation throughout the main paper and provide additional experiments on VGGT in the Appendix~\ref{app:vggt}.

\subsection{Geometry and Feature Distillation}
We train the masked student model by distilling geometric predictions from a frozen pretrained teacher. Given a multi-view sequence, the teacher is evaluated with dense global attention, while the student uses the masked global attention introduced above. We use the teacher predictions as supervision instead of sparse ground-truth annotations, and further validate this design choice through the ablations in Sec.~\ref{sec:ablate_distill} and analysis in the Appendix~\ref{app:sparse}.

\vspace{-10pt}
\paragraph{Geometry distillation.} Following previous works~\citep{wang2025vggt, wang2026pi}, the geometric loss consists of a scale-invariant point-map loss, a surface-normal loss, a confidence loss, and a relative camera-pose loss:
\begin{equation}
    \mathcal{L}_{\mathrm{geo}}
    =
    \mathcal{L}_{\mathrm{points}}
    +
    \lambda_{\mathrm{normal}}\mathcal{L}_{\mathrm{normal}}
    +
    \lambda_{\mathrm{conf}}\mathcal{L}_{\mathrm{conf}}
    +
    \lambda_{\mathrm{cam}}\mathcal{L}_{\mathrm{cam}} .
    \label{eq:geo_loss}
\end{equation}
We describe the point-map loss below and provide the definitions of the normal, confidence, and camera-pose losses in the Appendix~\ref{app:loss}.
Since the point maps are defined up to a scene-level scale, following ~\citet{wang2026pi}, we align the student prediction to the teacher pseudo target with a single scale factor $s^{*}$ . Let $\mathbf{p}^{S}_{i,j}$ and $\mathbf{p}^{T}_{i,j}$ denote the student and teacher 3D points at pixel $j$ of view $i$, respectively, and let $d^{T}_{i,j}$ denote the depth of the teacher point. The point-map distillation loss is then defined as
\begin{equation}
    \mathcal{L}_{\mathrm{points}}
    =
    \frac{1}{3NHW}
    \sum_{i=1}^{N}
    \sum_{j=1}^{HW}
    \frac{1}{d^{T}_{i,j}}
    \left\|
    s^{*}\mathbf{p}^{S}_{i,j}
    -
    \operatorname{sg}
    \left(
    \mathbf{p}^{T}_{i,j}
    \right)
    \right\|_{1},
    \label{eq:point_distill}
\end{equation}
where $\operatorname{sg}(\cdot)$ denotes stop-gradient.
\vspace{-10pt}

\paragraph{Feature distillation.}
In addition to output-level geometry supervision, we distill features before the prediction heads. Let $\mathbf{G}^{T,L}$ and $\mathbf{G}^{S,L}$ denote the final-layer features of the teacher and student. We split them into register tokens $\mathbf{R}$ and image tokens $\mathbf{Z}$,
\begin{equation}
    \mathbf{G}^{T,L}
    =
    [\mathbf{R}^{T};\mathbf{Z}^{T}],
    \qquad
    \mathbf{G}^{S,L}
    =
    [\mathbf{R}^{S};\mathbf{Z}^{S}],
\end{equation}
We apply cosine-distance losses to the two token groups:
\begin{equation}
    \mathcal{L}_{\mathrm{feat}}
    =
    \left(
    1
    -
    \operatorname{cos}
    \left(
    \mathbf{R}^{S},
    \operatorname{sg}
    \left(
    \mathbf{R}^{T}
    \right)
    \right)
    \right)
    +
    \lambda_{\mathrm{img}}
    \left(
    1
    -
    \operatorname{cos}
    \left(
    \mathbf{Z}^{S},
    \operatorname{sg}
    \left(
    \mathbf{Z}^{T}
    \right)
    \right)
    \right),
    \label{eq:feat_distill}
\end{equation}
where $\operatorname{cos}(\cdot,\cdot)$ denotes the mean cosine similarity over tokens.
The final training objective is
\begin{equation}
    \mathcal{L}
    =
    \mathcal{L}_{\mathrm{geo}}
    +
    \lambda_{\mathrm{feat}}
    \mathcal{L}_{\mathrm{feat}} .
    \label{eq:final_loss}
\end{equation}

\subsection{Anchor-Guided Adaptive Token Merging}
A key challenge for 3D foundation models is the quadratic cost of global attention, which limits scalability to long image sequences. Existing methods typically reduce this cost through token-level compression~\citep{shen2025fastvggt,wang2026flashvggt,shu2026litevggt}. However, token-level similarity does not necessarily imply geometric redundancy. Repetitive structures, such as windows on a building facade, may produce similar features and thus be merged into a single token despite corresponding to different 3D locations. Aggressive token merging can therefore remove spatially distinct geometric evidence. To better preserve scene structure, we introduce Anchor-Guided Adaptive token merging (AGA) to compress at the frame level instead of token level.


Given $N$ input frames, we uniformly sample a subset as anchor frames $\mathcal{A}$, which are kept at full token resolution and are distributed across the input sequence to provide a coarse but complete coverage. To measure redundancy, we mean-pool the DINOv2 tokens $\mathbf{X}_i$ into a frame descriptor and compute the maximum cosine similarity between each non-anchor frame and the anchor frames:
\begin{equation}
\mathbf{z}_i=\frac{1}{P}\sum_{p=1}^{P}\mathbf{X}_{i,p},
\qquad
s_i=\max_{a\in\mathcal{A}}
\frac{\mathbf{z}_i^\top\mathbf{z}_a}
{\lVert\mathbf{z}_i\rVert,\lVert\mathbf{z}_a\rVert},
\quad i\notin\mathcal{A}.
\end{equation}
A high value of $s_i$ indicates that frame $i$ is well represented by the anchor context and is likely to contain redundant information. Conversely, a low value of $s_i$ suggests that the frame provides complementary content and should preserve more tokens.
\vspace{-10pt}
\paragraph{Similarity-adaptive merging ratio.}
For each non-anchor frame $i\notin\mathcal{A}$, we map the redundancy score to a merging ratio as $r_i = k s_i + b$, where $k$ and $b$ are calibrated such that $\sum_{i\notin\mathcal{A}}\lfloor r_iP\rfloor = R_{\mathrm{target}}$ under the constraint $0\le r_i\le r_{\max}$.
Anchor frames use $r_a=0$ and remain uncompressed.

\vspace{-10pt}
\paragraph{Token merging and recovery.}
For each non-anchor frame $i\notin\mathcal{A}$, we apply ToMe-style bipartite matching~\citep{bolya2022token} to its unprotected patch tokens and merge $\lfloor r_iP\rfloor$ tokens, obtaining a compressed representation $\widetilde{\mathbf{F}}_i=\operatorname{Merge}(\mathbf{F}_i,r_i)$. Global attention is then computed over full-resolution anchor tokens and compressed non-anchor tokens during inference:
\begin{equation}
    \widetilde{\mathbf{G}}
    =
    \operatorname{SelfAttn}_{\mathrm{global}}
    \left(
    [
    \{\mathbf{F}_a\}_{a\in\mathcal{A}};
    \{\widetilde{\mathbf{F}}_i\}_{i\notin\mathcal{A}}
    ]
    \right).
\end{equation}
The merged non-anchor outputs are finally restored to the original token layout using the recorded merge assignments, i.e., $\mathbf{G}_i=\operatorname{Unmerge}(\widetilde{\mathbf{G}}_i)$, and are passed to the subsequent layers and prediction heads without architectural modification.

%% file: sec/4_experiments.tex
\section{Experiments}


\subsection{Training Datasets}
We distill our model on a large-scale multi-domain mixture following the data protocol of prior feed-forward 3D foundation models~\citep{zhuo2026streaming}. The training set consists of 13 datasets, including ARKitScenes~\citep{baruch2021arkitscenes}, BlendedMVS~\citep{yao2020blendedmvs}, Co3Dv2~\citep{reizenstein2021common}, HyperSim~\citep{roberts2021hypersim}, MegaDepth~\citep{li2018megadepth}, MVS-Synth~\citep{huang2018deepmvs}, OmniObject3D~\citep{wu2023omniobject3d}, PointOdyssey~\citep{zheng2023pointodyssey}, ScanNet~\citep{dai2017scannet}, Spring~\citep{mehl2023spring}, Virtual KITTI~\citep{cabon2020virtual}, Waymo~\citep{sun2020scalability}, and WildRGB~\citep{xia2024rgbd}. This mixture covers indoor and outdoor scenes, object-centric and scene-level captures, synthetic and real-world data, as well as both static and dynamic sequences. Such diversity is important for distilling robust geometric priors under masked cross-view connectivity. Details of model training can be found in
Appendix~\ref{app:implement}.
\begin{figure}
    \centering
    \includegraphics[width=1\linewidth]{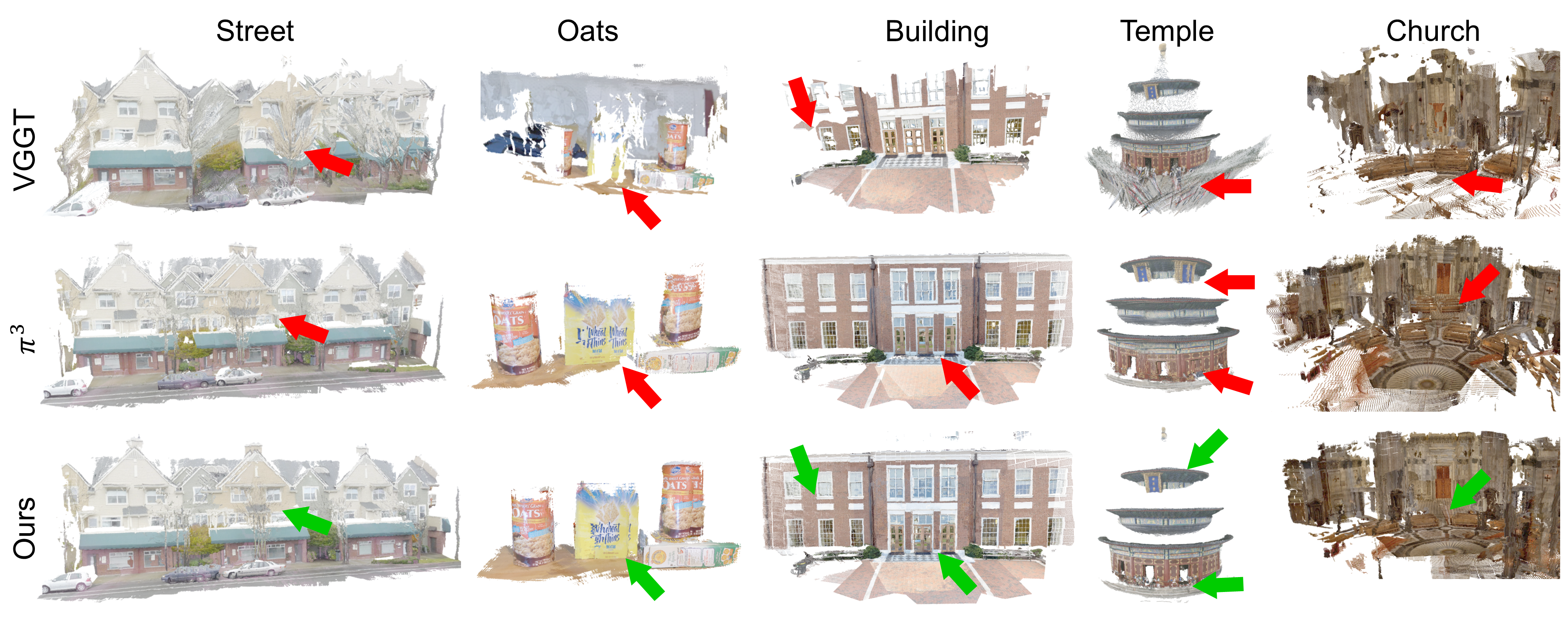}
    \caption{\textbf{Qualitative comparison of multi-view 3D reconstruction.} Compared with VGGT and $\pi^3$, our method produces more consistent reconstructions in standard scenes (Street, Oats) and in regions with repeated or visually similar structures, including building doors, temple signboards, and church benches, where the baselines often produce shifted or duplicated geometry.}
    \label{fig:vis}
    \vspace{-10pt}
\end{figure}
\subsection{Evaluation datasets and metrics}
We evaluate MGE on four aspects: robustness to visual ambiguity (Sec.~\ref{sec:doppel}), robustness to occlusion (Sec.~\ref{sec:eth3d_occlusion}), token merging performance (Sec.~\ref{sec:long_seq}), and standard reconstruction performance (Sec.~\ref{sec:standard}). Qualitative point-map visualizations are provided in Figure~\ref{fig:vis}, Figure~\ref{fig:vis_app} and Figure~\ref{fig:eth3d_qualitative}. We further conduct ablation studies to analyze the contribution of each component (Sec.~\ref{sec:ablation}). 

For 3D reconstruction task, predicted point maps are aligned to the ground truth using a similarity transformation followed by ICP. We report accuracy (Acc.), completeness (Comp.), chamfer distance (CD) and normal consistency (NC), where lower Acc./Comp./CD and higher NC are better. For camera trajectories, we report Sim(3)-aligned absolute trajectory error (ATE), relative translation error (RPE-t), and relative rotation error (RPE-r). For relative camera poses estimation, we report pose AUC@30, RRA@15 and RTA@15. Best and second-best results are shown in \textbf{bold} and \underline{underline}.

\subsection{Robustness to Visual Ambiguity}
\label{sec:doppel}

Visual ambiguities, i.e., regions with similar appearance but different geometric locations, often disrupt camera calibration. Since no official multi-view Doppelganger benchmark exists, we build a camera-pose benchmark under visual ambiguities from~\citet{heinly2014correcting}, using COLMAP reconstructions with Doppelgangers++~\citep{xiangli2025doppelgangers++} as pseudo ground truth. The benchmark contains 43 subsets, each consisting of 15 geometrically consistent reference views and 5 visually similar but geometrically inconsistent ``Doppelganger'' views. Each feed-forward model processes all 20 images jointly in a single forward pass. We also include LoFTR~\citep{sun2021loftr} as a traditional matching-based baseline, which estimates correspondences independently for each image pair and then performs geometric verification, triangulation, and global pose optimization. 

As shown in Table~\ref{tab:dopp}, LoFTR achieves the highest AUC@30 on the reference pairs. Since we use sequential pairwise matching, unrelated Doppelganger views have limited impact on the reconstruction of the reference images. However, some images fail to register, and the performance on reference--doppelganger cross pairs is much lower. Among the feed-forward methods, MGE achieves the best overall performance, with the strongest results on all-pair, reference-pair, and cross-pair evaluations. Qualitative camera-pose visualizations are provided in Appendix Figure~\ref{fig:cameras}. 
\input{tables/dpp}



\input{tables/robust_bold_underline_best}

\subsection{Robustness to Occlusion}
\label{sec:eth3d_occlusion}

We evaluate occlusion robustness on ETH3D~\citep{schops2017multi}, progressively corrupting 10\%/30\%/50\% views using a centered white patch to formulate small/medium/large occlusion scenarios. We evaluate dense reconstruction and camera pose under increasing levels of occlusion, considering both full-attention and token-merging methods.
Table~\ref{tab:occlusion} shows that our method performs best under medium and large occlusion. As occlusion increases from clean to large, VGGT and $\pi^3$ suffer 119\% and 128\% increases in accuracy error, respectively. Their fast variants show smaller relative degradation, but start from substantially worse reconstruction quality because of token merging and therefore remain less accurate overall. In contrast, our masked cross-view training improves representation robustness: the point accuracy error increases by only 48\%, while pose estimation even improves by 4\% under heavy occlusion. AGA further benefits from this robustness by retaining more tokens from less-occluded views, leading to a smaller performance drop as occlusion becomes more severe.
\vspace{-5pt}
\subsection{Token merging performance}
\label{sec:long_seq}
\begin{figure}
    \vspace{-10pt}
    \centering
    \includegraphics[width=1\linewidth]{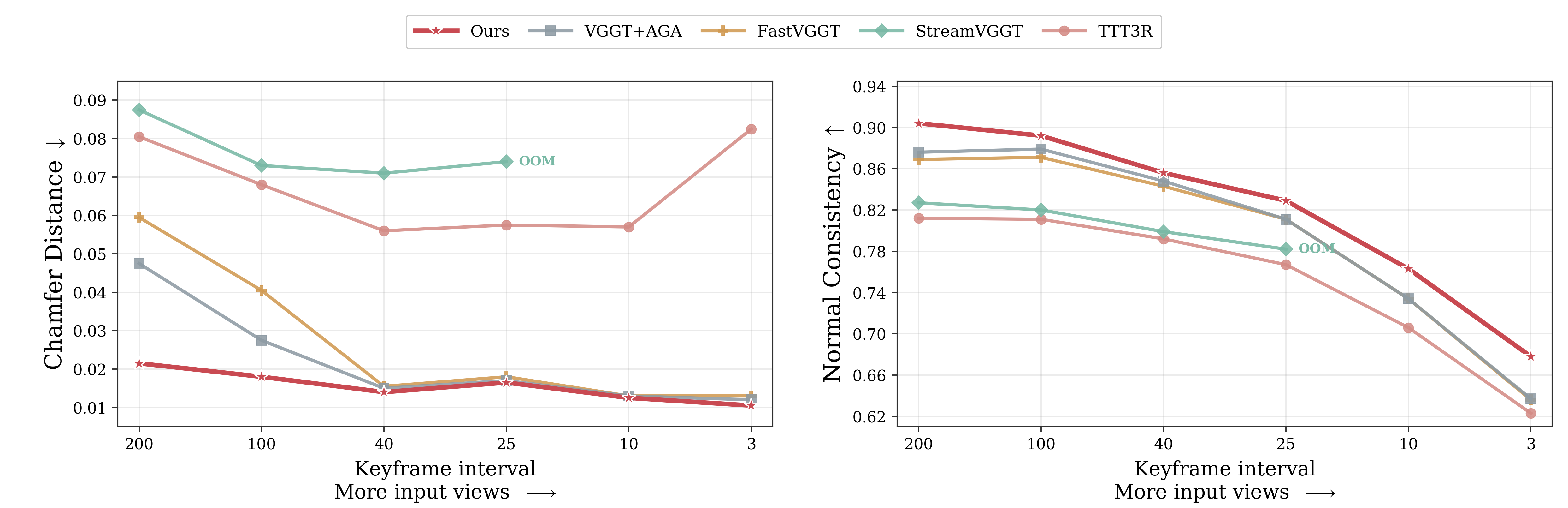}
    \caption{\textbf{Comparison of 3D Reconstruction on NRGBD.} Following FastVGGT~\citep{shen2025fastvggt}, we uniformly sample keyframes with strides from 200 to 3.}
    \label{fig:long_seq}
    \vspace{-20pt}
\end{figure}

Following FastVGGT~\citep{shen2025fastvggt}, we evaluate 3D reconstruction on NRGBD~\citep{azinovic2022neural} by measuring the distance between predicted point maps and the ground-truth point cloud. Figure~\ref{fig:long_seq} shows that our method consistently outperforms streaming approaches, including TTT3R~\citep{chen2026ttt3r} and StreamVGGT~\citep{zhuo2026streaming}, as well as existing token-merging method FastVGGT~\cite{shen2025fastvggt} across different sequence lengths. Applying AGA to the original VGGT already yields consistent gains over FastVGGT, while combining AGA with MGE further improves robustness, particularly in limited-view settings where aggressive token merging tends to discard important geometric evidence. Detailed quantitative results are provided in the Appendix~\ref{app:long_sequence}.



\input{tables/camera_pose}
\subsection{Standard Benchmarks}
\label{sec:standard}
We evaluate camera pose estimation on Sintel~\citep{NEURIPS2021_0a87257e}, TUM-Dynamics~\citep{sturm2012benchmark}, and ScanNet~\citep{dai2017scannet}, following the standard pose error metric used in prior work~\citep{zhang2025monst3r,wang2025continuous}. As shown in Table~\ref{tab:camera_pose}, the masked student remains highly competitive with the pretrained teacher across all datasets, matching its performance on TUM-Dynamics and slightly improving results on Sintel and ScanNet. The occasional gains further suggest that the proposed MGE can act as an effective regularizer rather than merely approximating the dense teacher. Further quantitative evaluations on additional benchmarks are provided in the Appendix~\ref{app:conv}.

\input{tables/ablation/maskratio}

\subsection{Ablation Study}
\label{sec:ablation}

\paragraph{Mask Designs.}
\label{sec:ablate_mask}
We ablate the masking ratio on NRGBD in Table~\ref{tab:ablation_mask}.The results reveal a trade-off between representation independence and reconstruction fidelity. Stronger masking encourages each frame to rely less on dense cross-view communication, but excessive sparsification removes useful geometric context and degrades reconstruction even when all tokens are retained at inference. Conversely, weaker masking preserves reconstruction quality but reduces the effectiveness of subsequent token merging on long sequences. In practice, moderate masking provides the best balance, with $\rho=0.3$--$0.5$. We further compare different masking patterns during training in the Appendix~\ref{app:ablate_pattern}.

\vspace{-10pt}

\paragraph{Merging Designs.}
\label{sec:ablate_merging}
\begin{wrapfigure}{r}{0.50\textwidth}
    \centering
    \includegraphics[width=0.50\textwidth]{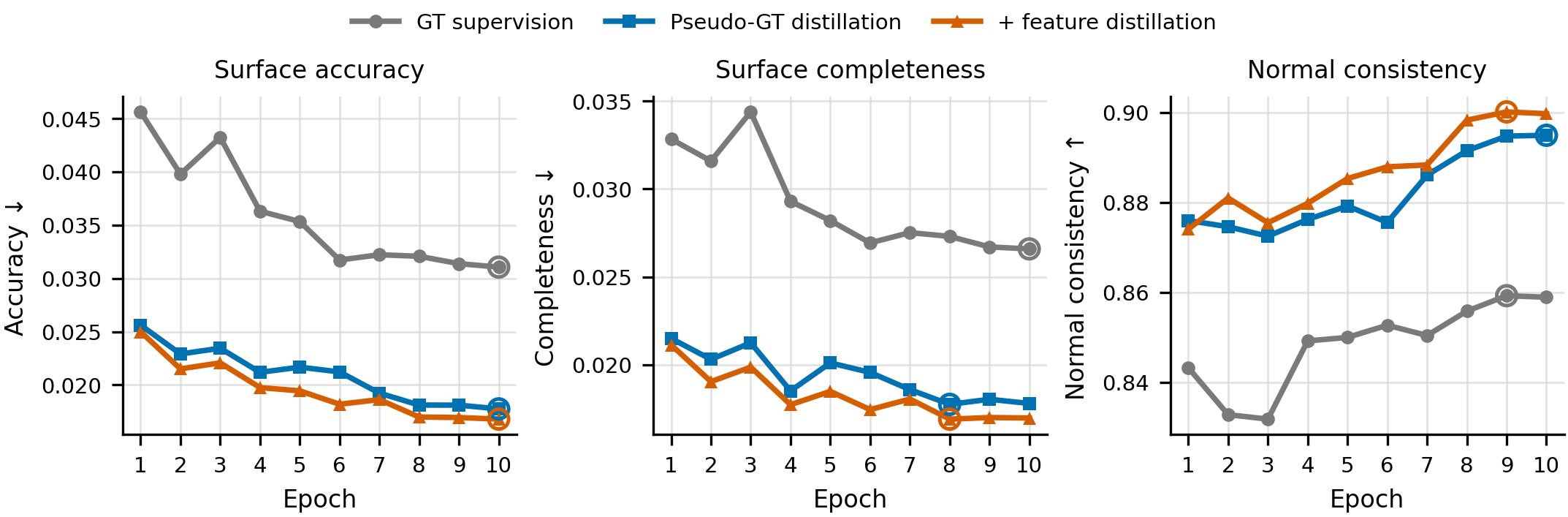}
    \caption{Ablation of various training strategies on NGRBD}
    \label{fig:distill_training_curves}
    \vspace{-10pt}
\end{wrapfigure}
We evaluate different merging designs on ETH3D~\citep{schops2017multi}, which contains many low-texture and repetitive regions. Starting from the FastVGGT-style baseline, we replace uniformly protected tokens with uniformly sampled anchor frames, and then further introduce adaptive frame-wise merging. All variants retain the same overall token ratio for a fair comparison. As shown in Table~\ref{tab:ablation_merging}, anchor-frame protection substantially improves reconstruction quality over token-level protection, while adaptive merging provides additional gains. Comparing AGA to both VGGT and MGE weights, the marginal gains show that the representations learned by MGE are more robust to aggressive frame-level compression.

\vspace{-10pt}
\paragraph{Distillation Strategies.}
\label{sec:ablate_distill}
To assess the effectiveness of our training strategy, we compare direct ground-truth supervision, output-level pseudo supervision, and additional feature supervision on NRGBD. All variants are trained with the same optimization budget, and intermediate checkpoints are evaluated to compare convergence. Figure~\ref{fig:distill_training_curves} shows that direct supervision converges more slowly and yields consistently higher reconstruction error. Pseudo supervision improves both convergence and reconstruction quality, while feature supervision further narrows the gap to the teacher.

%% file: tables/dpp.tex
\begin{table}[htp]
    \vspace{-0.5em}
    \centering
    \caption{
        \textbf{Visual ambiguity evaluation.} We report relative-pose accuracy over all image pairs, reference-only pairs, and reference--doppelganger cross pairs. Higher is better.}
    \vspace{-1em}
    \resizebox{\columnwidth}!{
    \begin{tabular}{lccccccccc}
        \toprule[0.17em]
        {\multirow{3}{*}{\textbf{Method}}} &  \multicolumn{3}{c}{\textbf{All}} & \multicolumn{3}{c}{\textbf{Reference}} & \multicolumn{3}{c}{\textbf{Cross}} \\
      \cmidrule(r){2-4} \cmidrule(r){5-7}  \cmidrule(r){8-10}
   & AUC@30 & RRA@15 & RTA@15	& AUC@30  & RRA@15 & RTA@15 & AUC@30  & RRA@15 & RTA@15 \\
    \midrule[0.08em]
      LoFTR & \textbf{59.60}&60.34 &64.68 &\textbf{88.22}& 89.13 & \textbf{91.30}  & 16.86 & 16.90& 24.78  \\
      VGGT &55.56 &\underline{76.40} & \underline{65.95}&65.75& \underline{93.47} & 75.90 & \underline{42.21}  & \underline{50.95} & \underline{53.95} \\
    $\pi^3$ &51.23&70.73&62.48 & 59.42 &85.40 & 69.50  & 39.10 & 47.41 & 52.34\\
    \textbf{MGE}& \underline{58.68} &\textbf{77.52} & \textbf{70.07} & \underline{70.51}& \textbf{94.99} &  \underline{81.48} & \textbf{42.72} & \textbf{51.60} & \textbf{54.60}\\
    \bottomrule[0.17em] 
    \end{tabular}
    }
    \vspace{-1.5em}
    \label{tab:dopp}
\end{table}

%% file: tables/robust_bold_underline_best.tex
\begin{table}[htp]
    \vspace{-0.5em}
    \centering
    \caption{
        \textbf{Occlusion evaluation on ETH3D.} 
    }
    \vspace{-1em}
    \resizebox{1.0\columnwidth}!{
    \begin{tabular}{lccccccccccccc}
        \toprule[0.17em]
      \textbf{Condition} 
& \multicolumn{3}{c}{Clean}
& \multicolumn{3}{c}{Small}
& \multicolumn{3}{c}{Medium}
& \multicolumn{3}{c}{Large} \\
\cmidrule(r){1-1}\cmidrule(r){2-4} \cmidrule(r){5-7} \cmidrule(r){8-10}
        \cmidrule(r){11-13} 
       Methods&Acc. $\downarrow$ & Comp. $\downarrow$  & NC. $\uparrow$ & Acc. $\downarrow$ & Comp. $\downarrow$  & NC. $\uparrow$& Acc. $\downarrow$ & Comp. $\downarrow$  & NC. $\uparrow$ & Acc. $\downarrow$ & Comp. $\downarrow$  & NC. $\uparrow$  \\
       
        \midrule[0.08em]
        VGGT                   & 0.149   & 0.124  & 0.849  & 0.178   & 0.133  & 0.841  & 0.268   & 0.168   & 0.818  & 0.327   & 0.224  & 0.799  \\

$\pi^3$                 & \underline{0.122}   & \textbf{0.092}  & \textbf{0.872}  & \textbf{0.114}   & \textbf{0.092}  & \textbf{0.871}  & \underline{0.181}   & \underline{0.144}   & \underline{0.855}  & \underline{0.278}   & \underline{0.195}  & \underline{0.823}  \\
\textbf{MGE}            & \textbf{0.116}   & \underline{0.101}  & \underline{0.871}  & \underline{0.121}   & \underline{0.096}  & \underline{0.863}  & \textbf{0.142}   & \textbf{0.107}   & \textbf{0.860}  & \textbf{0.171}   & \textbf{0.129}  & \textbf{0.847} \\
\midrule[0.08em]
FastVGGT & 0.313 & 0.351 & 0.800 & 0.321 & 0.281 & 0.791 & 0.383 & 0.359 & 0.783 & 0.423 & 0.314 & 0.767 \\
Fast-$\pi^3$ & \underline{0.163} & \underline{0.141} & \underline{0.837} & \underline{0.184} & \underline{0.156} & \underline{0.824} & \underline{0.188} & \underline{0.155} & \textbf{0.829} & \underline{0.225} & \underline{0.172} & \underline{0.816} \\
\textbf{MGE+AGA} & \textbf{0.156} & \textbf{0.138} & \textbf{0.838} & \textbf{0.154} & \textbf{0.126} & \textbf{0.832} & \textbf{0.175} & \textbf{0.136} & \underline{0.823} & \textbf{0.182} & \textbf{0.143} & \textbf{0.819} \\ 
        \midrule[0.17em]
        Methods & ATE $\downarrow$ & RPE-t $\downarrow$ & RPE-r $\downarrow$ &
        ATE $\downarrow$ & RPE-t $\downarrow$ & RPE-r $\downarrow$ &
        ATE $\downarrow$ & RPE-t $\downarrow$ & RPE-r $\downarrow$ &
        ATE $\downarrow$ & RPE-t $\downarrow$ & RPE-r $\downarrow$ \\
        \midrule[0.08em]
        VGGT           & 0.716   & 0.887  & 2.122  & 0.738   & 0.941  & 2.155  & 0.758   & 0.961   & \underline{2.948}  & 0.898   & 1.227  & \underline{4.673}  \\
        

$\pi^3$                 & \underline{0.248}   & \underline{0.334}  & \underline{2.047}  & \textbf{0.155}   & \textbf{0.230}  & \underline{1.679} & \underline{0.264}   & \underline{0.355}   & 4.502  & \underline{0.361}   & \underline{0.458}  & 7.969  \\

\textbf{MGE}                & \textbf{0.214}   & \textbf{0.319}  & \textbf{1.750}  & \underline{0.159}   & \underline{0.250}  & \textbf{1.614}  & \textbf{0.199}   & \textbf{0.286}   & \textbf{2.140}  & \textbf{0.205}   & \textbf{0.293}  & \textbf{2.043}
\\
\midrule[0.08em]
FastVGGT & 1.178 & 1.529 & 8.490 & 1.147 & 1.486 & 8.085 & 1.213 & 1.552 & 10.537 & 1.290 & 1.659 & 10.699          &   \\
Fast-$\pi^3$ & \underline{0.499} & \underline{0.668} & \underline{3.140} & \underline{0.542} & \underline{0.711} & \underline{3.170} & \underline{0.587} & \underline{0.765} & \underline{3.894} & \underline{0.556} & \underline{0.774} & \underline{5.034} \\
\textbf{MGE+AGA} & \textbf{0.448} & \textbf{0.602} & \textbf{2.379} & \textbf{0.432} & \textbf{0.561} & \textbf{2.526} & \textbf{0.456} & \textbf{0.631} & \textbf{2.819} & \textbf{0.427} & \textbf{0.589} & \textbf{2.798} \\
        \bottomrule[0.17em] 
    \end{tabular}
    }
    \label{tab:occlusion}
\end{table}

%% file: tables/camera_pose.tex
\begin{table}[t]
    \vspace{-1em}
    \centering
    \caption{\textbf{Camera pose estimation over standard benchmarks.}
    }
    \vspace{-1em}
    \resizebox{1.0\columnwidth}!{
    \begin{tabular}{lccccccccc}
        \toprule[0.17em]
        \multicolumn{1}{l}{\multirow{3}{*}{\textbf{Method}}} &
        \multicolumn{3}{c}{\textbf{Sintel}} &
        \multicolumn{3}{c}{\textbf{TUM-dynamics}} &
        \multicolumn{3}{c}{\textbf{ScanNet}~(seen)}\\
        \cmidrule(r){2-4} \cmidrule(r){5-7} \cmidrule(r){8-10}
        \multicolumn{1}{c}{} &
        ATE $\downarrow$ & RPE-t $\downarrow$ & RPE-r $\downarrow$ &
        ATE $\downarrow$ & RPE-t $\downarrow$ & RPE-r $\downarrow$ &
        ATE $\downarrow$ & RPE-t $\downarrow$ & RPE-r $\downarrow$  \\
        \midrule[0.08em] 
        Fast3R~\citep{yang2025fast3r}  & 0.371 & 0.298 & 13.75 & 0.090 & 0.101 & 1.425 & 0.155 & 0.123 & 3.491 \\
        CUT3R~\citep{wang2025continuous} & 0.217 & 0.070 & 0.636 & 0.047 & 0.015 & 0.451 & 0.094 & 0.022 & 0.629 \\
        FLARE~\citep{zhang2025flare} & 0.207 & 0.090 & 3.015 & 0.026 & 0.013 & 0.475 & 0.064 & 0.023 & 0.971 \\
        VGGT~\citep{wang2025vggt}  & 0.167 & 0.062 & 0.491 & \underline{0.012} & 0.010 & 0.311 & 0.035 & 0.015 & 0.382  \\
        $\pi^3$~\citep{wang2026pi} & \underline{0.074} & \underline{0.040} & \underline{0.282} & 0.014 & 0.009 & 0.312 & \underline{0.031} & \underline{0.013} & \textbf{0.347}  \\
        \textbf{MGE} & \textbf{0.071} & \textbf{0.033} & \textbf{0.234} & 0.014 & \underline{0.009} & \textbf{0.306} & \textbf{0.029} & \textbf{0.012} & \underline{0.348} \\
        \bottomrule[0.17em]
    \end{tabular}
    }
    \vspace{-0.5em}
    \label{tab:camera_pose}
\end{table}

%% file: tables/ablation/maskratio.tex
\begin{figure}[t]
    \centering

    \begin{minipage}[t]{0.4\linewidth}
        \centering
        \captionof{table}{\textbf{Effect of mask ratio $\rho$.} We compare different masking ratios under two keyframe sampling strides.}
        \vspace{-0.5em}
        \resizebox{\linewidth}{!}{
        \begin{tabular}{lcccc}
            \toprule
            \multirow{2}{*}{Setting}
            & \multicolumn{2}{c}{\textbf{stride=100}}
            & \multicolumn{2}{c}{\textbf{stride=10}} \\
            \cmidrule(lr){2-3} \cmidrule(lr){4-5}
            & CD $\downarrow$ & NC $\uparrow$
            & CD $\downarrow$ & NC $\uparrow$ \\
            \midrule
            $\rho=0.3$ & \textbf{0.0150} & \textbf{0.9184}
            & \underline{0.0134} & \underline{0.7607} \\
            $\rho=0.5$ & \underline{0.0159} & \underline{0.9104}
            & \textbf{0.0126} & \textbf{0.7660} \\
            $\rho=0.7$ & 0.0182 & 0.8983
            & 0.0156 & 0.7440 \\
            \bottomrule
        \end{tabular}
        }
        \label{tab:ablation_mask}
    \end{minipage}
    \hfill
    \begin{minipage}[t]{0.56\linewidth}
        \centering
        \captionof{table}{\textbf{Ablation of merging designs on ETH3D.}
        All variants retain the same overall token ratio for a fair comparison.}
        \vspace{-0.5em}
        \resizebox{\linewidth}{!}{
        \begin{tabular}{lccc}
            \toprule
            Method & Acc. $\downarrow$ & Comp. $\downarrow$ & NC $\uparrow$ \\
            \midrule
            MGE+FastVGGT-style merging
            & 0.308 & 0.311 & 0.815 \\
            MGE+Replace with anchor frames
            & 0.230 & 0.256 & 0.838 \\
            VGGT+AGA & 0.305 & 0.311 & 0.821 \\
            MGE+AGA
            & \textbf{0.216} & \textbf{0.243} & \textbf{0.842} \\
            
            \bottomrule
        \end{tabular}
        }
        \label{tab:ablation_merging}
    \end{minipage}

    \vspace{-1.2em}
\end{figure}

%% file: sec/5_conclusion.tex
\section{Conclusion}
In this paper, we presented a masked geometric encoder for improving the robustness and efficiency of feed-forward 3D reconstruction. By randomly masking frame-to-frame connections during fine-tuning, the model is encouraged to recover consistent geometry from incomplete cross-view communication and to rely less on individual, potentially misleading interactions. A teacher-guided distillation strategy preserves the geometric capability of the pretrained model while enabling more efficient optimization of the masked student. At inference time, we further exploit the resulting frame-level redundancy through anchor-guided adaptive token merging, preserving complete anchor views while allocating token budgets according to cross-view redundancy.

Extensive experiments show that MGE preserves dense-teacher reconstruction quality while improving robustness to occlusion and visually ambiguous Doppelganger views. AGA further improves long-sequence efficiency, especially in low-texture and repetitive scenes. Together, these results demonstrate that regularizing cross-view connectivity is an effective way to improve the robustness and scalability of feed-forward 3D foundation models.

%% file: sec/appendix.tex
\section{Appendix}
\subsection{Dense Pseudo-Ground-Truth Supervision}
\label{app:sparse}
\begin{figure}[t]
    \centering
    \includegraphics[width=0.5\linewidth]{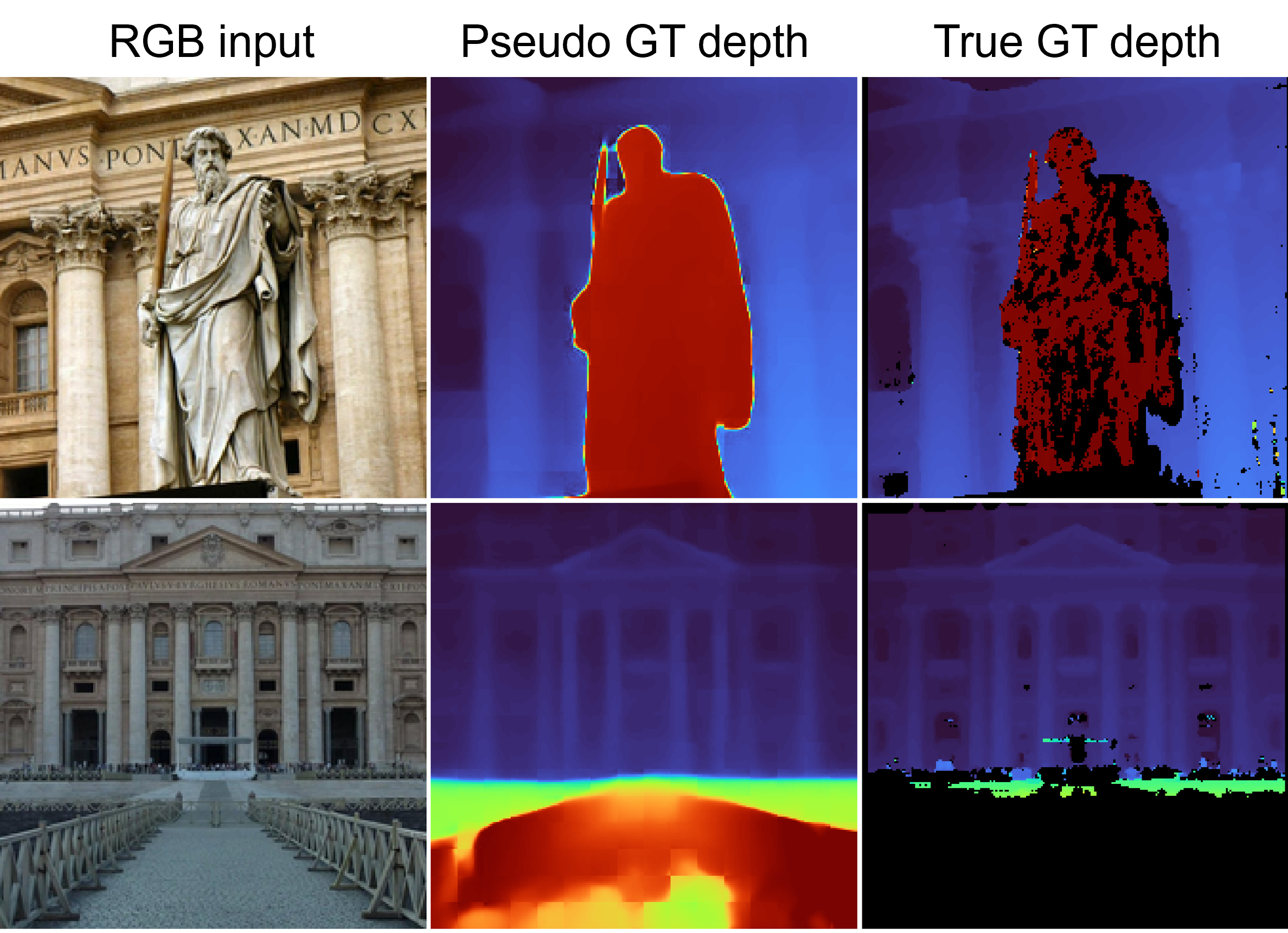}
    \caption{Qualitative comparison of supervision signals on MegaDepth. From left to right: RGB input, scale-aligned $\pi^3$ pseudo-GT depth, and dataset GT depth. $\pi^3$ supplies dense, spatially coherent geometry where the original annotations are incomplete; black pixels indicate missing GT.}
    \label{fig:gt_pseudo}
\end{figure}

Fig.~\ref{fig:gt_pseudo} qualitatively compares the input RGB images (left), dense pseudo-ground truth predicted by the pretrained teacher (middle), and the available ground-truth depth maps (right) from  MegaDepth~\citep{li2018megadepth}. For visualization, the pseudo-depth is aligned to the ground-truth scale independently for each view, and both depth maps use the same color range. The MegaDepth annotations contain missing and invalid regions, shown in black, whereas the pretrained 3D feed-forward model produces spatially dense depth estimates over the complete image. Despite being dense, the pseudo-ground truth preserves the principal scene structure, object boundaries, and relative depth ordering observed in the available annotations, which encourages us to use it as supervision.

\subsection{Training loss}
\label{app:loss}
We train the masked student by distilling the geometric predictions of a pretrained $\pi^3$ teacher. Given a multi-view sequence $\mathcal{I}=\{I_i\}_{i=1}^{N}$, the frozen teacher is evaluated with dense global attention, while the student uses the masked global attention. Following the $\pi^3$ output formulation, the prediction heads decode the final per-frame representations into camera poses, local point maps, and confidence maps:
\begin{equation}
    \left(
    \mathbf{T}^{T}_{i},
    \mathbf{P}^{T}_{i},
    \mathbf{C}^{T}_{i}
    \right)
    =
    h_{\theta_T}(\mathbf{G}^{T}_{i}),
    \qquad
    \left(
    \mathbf{T}^{S}_{i},
    \mathbf{P}^{S}_{i},
    \mathbf{C}^{S}_{i}
    \right)
    =
    h_{\theta_S}(\mathbf{G}^{S}_{i}),
    \qquad i=1,\ldots,N.
\end{equation}
Here, $\mathbf{T}_{i}\in SE(3)$ denotes the camera pose, $\mathbf{P}_{i}\in\mathbb{R}^{H\times W\times 3}$ denotes the pixel-aligned local point map in the coordinate system of view $i$, and $\mathbf{C}_{i}\in\mathbb{R}^{H\times W}$ denotes the confidence map.

\paragraph{Normal distillation.}
Following $\pi^3$, we further supervise local surface geometry with a normal loss. Let $\mathbf{n}^{S}_{i,j}$ and $\mathbf{n}^{T}_{i,j}$ denote the surface normals computed from the student and teacher point maps, respectively. We define
\begin{equation}
    \mathcal{L}_{\mathrm{normal}}
    =
    \frac{1}{NHW}
    \sum_{i=1}^{N}
    \sum_{j=1}^{HW}
    \arccos
    \left(
    \mathbf{n}^{S}_{i,j}
    \cdot
    \operatorname{sg}
    \left(
    \mathbf{n}^{T}_{i,j}
    \right)
    \right).
\end{equation}

\paragraph{Confidence distillation.}
We supervise the student confidence map using a binary target derived from the pseudo geometric error. Specifically, the target confidence is set to one if the scale-aligned point-map error is below a threshold $\epsilon$, and zero otherwise:
\begin{equation}
    y_{i,j}
    =
    \mathbbm{1}
    \left[
    \frac{1}{z^{T}_{i,j}}
    \left\|
    s^{*}\mathbf{x}^{S}_{i,j}
    -
    \operatorname{sg}
    \left(
    \mathbf{x}^{T}_{i,j}
    \right)
    \right\|_{1}
    <
    \epsilon
    \right].
\end{equation}
The confidence loss is
\begin{equation}
    \mathcal{L}_{\mathrm{conf}}
    =
    \frac{1}{NHW}
    \sum_{i=1}^{N}
    \sum_{j=1}^{HW}
    \operatorname{BCE}
    \left(
    \mathbf{C}^{S}_{i,j},
    y_{i,j}
    \right).
\end{equation}

\paragraph{Relative camera-pose distillation.}
Because the predicted camera poses are defined up to a global similarity transformation, we supervise the student using relative poses between views. For each ordered pair $i\neq j$, we compute
\begin{equation}
    \widehat{\mathbf{T}}^{S}_{i\leftarrow j}
    =
    \left(
    \mathbf{T}^{S}_{i}
    \right)^{-1}
    \mathbf{T}^{S}_{j},
    \qquad
    \widehat{\mathbf{T}}^{T}_{i\leftarrow j}
    =
    \left(
    \mathbf{T}^{T}_{i}
    \right)^{-1}
    \mathbf{T}^{T}_{j}.
\end{equation}
Let $\widehat{\mathbf{R}}^{S}_{i\leftarrow j}$ and $\widehat{\mathbf{t}}^{S}_{i\leftarrow j}$ denote the relative rotation and translation of the student, and let $\widehat{\mathbf{R}}^{T}_{i\leftarrow j}$ and $\widehat{\mathbf{t}}^{T}_{i\leftarrow j}$ denote the corresponding teacher targets. The rotation loss is
\begin{equation}
    \mathcal{L}_{\mathrm{rot}}(i,j)
    =
    \arccos
    \left(
    \frac{
    \operatorname{Tr}
    \left(
    \left(
    \operatorname{sg}
    \left(
    \widehat{\mathbf{R}}^{T}_{i\leftarrow j}
    \right)
    \right)^{\top}
    \widehat{\mathbf{R}}^{S}_{i\leftarrow j}
    \right)
    -1
    }{2}
    \right).
\end{equation}
The translation loss uses the same scale factor $s^{*}$ as the point-map alignment:
\begin{equation}
    \mathcal{L}_{\mathrm{trans}}(i,j)
    =
    H_{\delta}
    \left(
    s^{*}
    \widehat{\mathbf{t}}^{S}_{i\leftarrow j}
    -
    \operatorname{sg}
    \left(
    \widehat{\mathbf{t}}^{T}_{i\leftarrow j}
    \right)
    \right),
\end{equation}
where $H_{\delta}(\cdot)$ denotes the Huber loss. The camera distillation loss is
\begin{equation}
    \mathcal{L}_{\mathrm{cam}}
    =
    \frac{1}{N(N-1)}
    \sum_{i\neq j}
    \left(
    \mathcal{L}_{\mathrm{rot}}(i,j)
    +
    \lambda_{\mathrm{trans}}
    \mathcal{L}_{\mathrm{trans}}(i,j)
    \right).
\end{equation}

$\mathcal{L}_{\mathrm{points}}$ is defined as Eq.~\ref{eq:point_distill}. The overall training objective is
\begin{equation}
    \mathcal{L}
    =
    \mathcal{L}_{\mathrm{points}}
    +
    \lambda_{\mathrm{normal}}
    \mathcal{L}_{\mathrm{normal}}
    +
    \lambda_{\mathrm{conf}}
    \mathcal{L}_{\mathrm{conf}}
    +
    \lambda_{\mathrm{cam}}
    \mathcal{L}_{\mathrm{cam}}
    +
    \lambda_{\mathrm{feat}}
    \mathcal{L}_{\mathrm{feat}}.
\end{equation}

\begin{figure}
    \centering
    \includegraphics[width=1\linewidth]{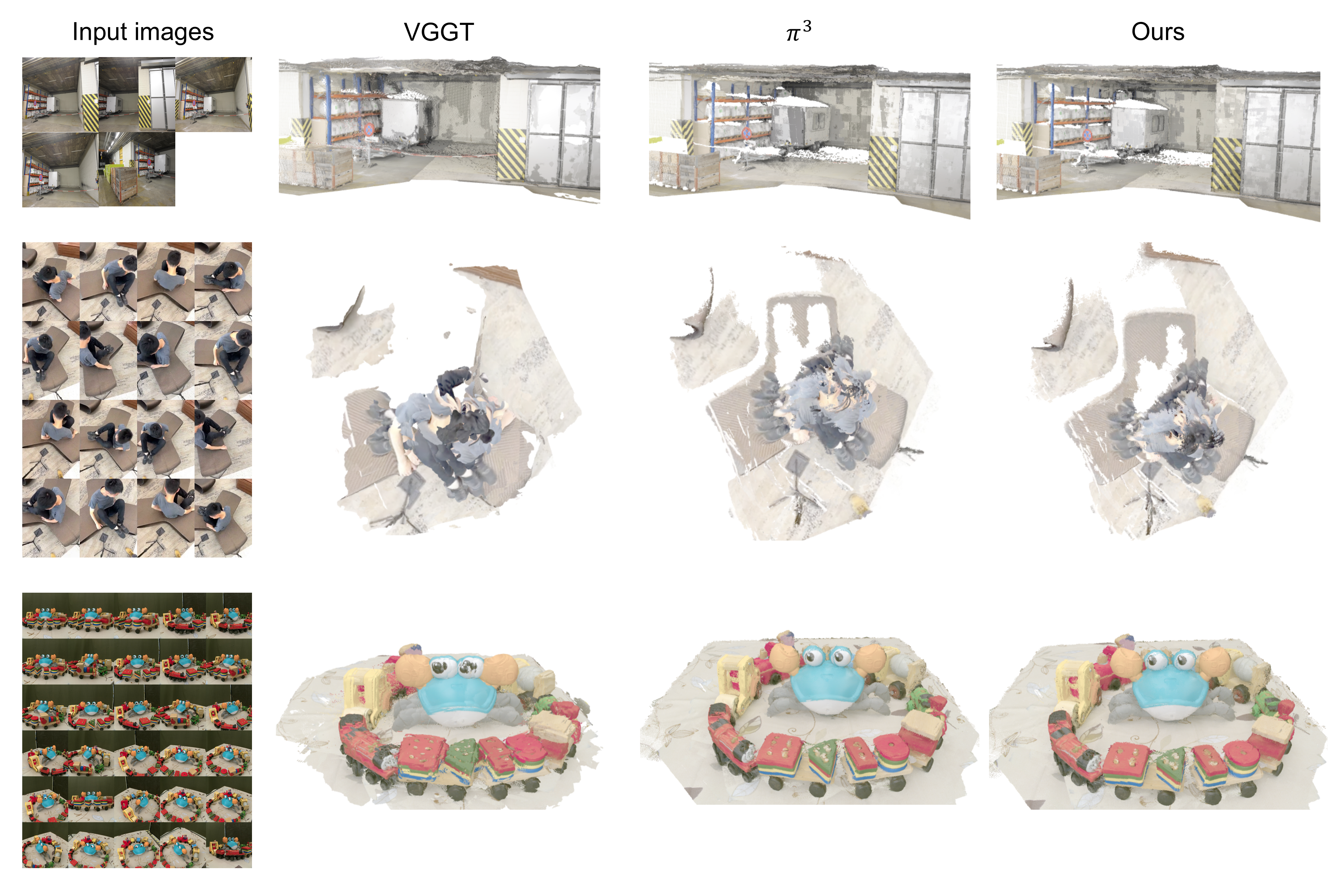}
    \caption{More qualitative comparison of multi-view 3D reconstruction. The second row highlights a representative occlusion case, where a person moves over the bench and causes severe cross-view occlusions. Our method better recovers the occluded part and preserves its geometry.}
    \label{fig:vis_app}
\end{figure}
\subsection{Implementation Details}
\label{app:implement}
We initialize the weights for the encoder and the alternating attention module from the pretrained teacher model, and we keep the encoder frozen during training. For our loss function, we set the weights for
each component as follows: $\lambda_{\mathrm{normal}} = 1.0,  \lambda_{\mathrm{cam}} = 0.1,  \lambda_{\mathrm{feat}}=0.1, \lambda_{\mathrm{img}}=100$. We employed the
AdamW optimizer and a hybrid schedule of linear warm-up (first 0.5 epochs) followed by cosine decay, reaching a peak learning rate of 1e-6. For each training iteration, we randomly sampled a batch of 24 frames from diverse training scenes. Following Point3R~\citep{wu2026point3r}, we processed
input images with variable aspect ratios while resizing the maximum edge length to 518 pixels. Training was conducted on 4 NVIDIA H200 GPUs for 6 days. We train the confidence head separately. After completing the main training stages, we freeze the rest of the network and train the confidence head in isolation by setting $\lambda_{\mathrm{conf}} = 0.05$. This stage converges rapidly and has no impact on the model’s overall performance.

\subsection{Generalization to VGGT.}
\label{app:vggt}
Our masked training framework is not specific to $\pi^3$ and can be applied to other
feed-forward 3D reconstruction models. We further validate this by using
VGGT~\citep{wang2025vggt} as the teacher model. Unlike $\pi^3$, VGGT represents the scene
in the coordinate system of the first input camera and uses the first frame as the
reference view. We therefore preserve the connection from every frame to the first
frame when sampling the random attention mask, while masking the remaining
cross-frame interactions in the same way as our default setting.

Following the output formulation of VGGT~\citep{wang2025vggt}, the teacher predicts camera
parameters $\mathbf{g}_i^T$, depth maps $\mathbf{D}_i^T$, and point maps
$\mathbf{P}_i^T$, which are used as pseudo supervision for the masked student.
The geometry distillation objective is
\begin{equation}
    \mathcal{L}_{\mathrm{geo}}
    =
    \mathcal{L}_{\mathrm{cam}}
    + \mathcal{L}_{\mathrm{depth}}
    + \mathcal{L}_{\mathrm{pmap}} .
\end{equation}
For camera prediction, we follow VGGT and use a Huber loss,
\begin{equation}
    \mathcal{L}_{\mathrm{cam}}
    =
    \sum_{i=1}^{N}
    \left\|
        \mathbf{g}_i^S-\operatorname{sg}(\mathbf{g}_i^T)
    \right\|_{\epsilon},
\end{equation}
where $\operatorname{sg}(\cdot)$ denotes stop-gradient on the teacher prediction.
For depth and point maps, we follow the uncertainty-weighted formulation used by
VGGT:
\begin{align}
    \mathcal{L}_{\mathrm{depth}}
    =&
    \sum_{i=1}^{N}
    \left\|
        \mathbf{\Sigma}^{D}_i
        \odot
        (\mathbf{D}_i^S-\mathbf{D}_i^T)
    \right\|_1
    +
    \left\|
        \mathbf{\Sigma}^{D}_i
        \odot
        (\nabla\mathbf{D}_i^S-\nabla\mathbf{D}_i^T)
    \right\|_1
    - \alpha \log \mathbf{\Sigma}^{D}_i ,
    \\
    \mathcal{L}_{\mathrm{pmap}}
    =&
    \sum_{i=1}^{N}
    \left\|
        \mathbf{\Sigma}^{P}_i
        \odot
        (\mathbf{P}_i^S-\mathbf{P}_i^T)
    \right\|_1
    +
    \left\|
        \mathbf{\Sigma}^{P}_i
        \odot
        (\nabla\mathbf{P}_i^S-\nabla\mathbf{P}_i^T)
    \right\|_1
    - \alpha \log \mathbf{\Sigma}^{P}_i .
\end{align}
We use the same feature distillation as Eq.~\ref{eq:feat_distill}.
Therefore, the overall training objective is
\begin{equation}
    \mathcal{L}
    =
    \mathcal{L}_{\mathrm{pmap}}
    +
    \mathcal{L}_{\mathrm{depth}}
    +
    \lambda_{\mathrm{cam}}
    \mathcal{L}_{\mathrm{cam}}
    +
    \lambda_{\mathrm{feat}}
    \mathcal{L}_{\mathrm{feat}}.
\end{equation}

We refer to the resulting model as MGE-VGGT and evaluate it on the ETH3D occlusion benchmark. As shown in Table~\ref{tab:append_occ}, MGE-VGGT remains close to the original VGGT under the clean setting and becomes increasingly better as the occlusion level increases. For dense reconstruction, VGGT's accuracy error increases from 0.149 to 0.327 from clean to large occlusion, whereas MGE-VGGT increases from 0.158 to 0.287. A similar trend is observed for completeness.

The difference is more pronounced for camera pose. Under large occlusion, MGE-VGGT reduces ATE from 0.898 to 0.805 and RPE-r from 4.673 to 2.111 compared with VGGT. While the two models perform similarly under clean and small occlusion, VGGT degrades rapidly under medium and large occlusion, whereas MGE-VGGT remains relatively stable. These results show that the robustness brought by masked cross-view training also transfers to VGGT, despite its different architecture and prediction heads, suggesting that MGE is not specific to our default $\pi^3$ teacher. 

We also evaluated MGE-VGGT on other standard benchmarks, which are reported in Appendix~\ref{app:conv}.

\begin{table}[htp]
    \centering
    \caption{
        \textbf{Occlusion evaluation on ETH3D for VGGT.}
    }
    \resizebox{1.0\columnwidth}!{
    \begin{tabular}{lccccccccccccc}
        \toprule[0.17em]
      \textbf{Condition} 
& \multicolumn{3}{c}{Clean}
& \multicolumn{3}{c}{Small}
& \multicolumn{3}{c}{Medium}
& \multicolumn{3}{c}{Large} \\
\cmidrule(r){1-1}\cmidrule(r){2-4} \cmidrule(r){5-7} \cmidrule(r){8-10}
        \cmidrule(r){11-13} 
       Methods&Acc. $\downarrow$ & Comp. $\downarrow$  & NC. $\uparrow$ & Acc. $\downarrow$ & Comp. $\downarrow$  & NC. $\uparrow$& Acc. $\downarrow$ & Comp. $\downarrow$  & NC. $\uparrow$ & Acc. $\downarrow$ & Comp. $\downarrow$  & NC. $\uparrow$  \\
       
        \midrule[0.08em]
        VGGT                   & \textbf{0.149}   & \textbf{0.124}  & \textbf{0.849}  & 0.178   & 0.133  & \textbf{0.841}  & 0.268   & 0.168   & \textbf{0.818}  & 0.327   & 0.224  &\textbf{0.799}  \\

        \textbf{MGE-VGGT}                  & 0.158 & 0.132 & 0.840 &\textbf{0.175} & \textbf{0.128} & 0.836& \textbf{0.252} & \textbf{0.160} &0.808 & \textbf{0.287} & \textbf{0.185} &  0.792 \\
        \midrule[0.17em]
        Methods & ATE $\downarrow$ & RPE-t $\downarrow$ & RPE-r $\downarrow$ &
        ATE $\downarrow$ & RPE-t $\downarrow$ & RPE-r $\downarrow$ &
        ATE $\downarrow$ & RPE-t $\downarrow$ & RPE-r $\downarrow$ &
        ATE $\downarrow$ & RPE-t $\downarrow$ & RPE-r $\downarrow$ \\
        \midrule[0.08em]
        VGGT           &\textbf{0.716}   & 0.887  & 2.122  & 0.738   & 0.941  & 2.155  & 0.758   & 0.961   & 2.948  & 0.898   & 1.227  & 4.673  \\
        
        \textbf{MGE-VGGT} &   0.723 & \textbf{0.885} & \textbf{1.790} & \textbf{0.733} & \textbf{0.899} & \textbf{1.774} & \textbf{0.757} & \textbf{0.926} & \textbf{1.936} & \textbf{0.805} & \textbf{0.995} & \textbf{2.111}           \\
        \bottomrule[0.17em] 
    \end{tabular}
    }
    \label{tab:append_occ}
\end{table}


\subsection{Visual Ambiguity Evaluation.}
\label{app:dopp}
\subsubsection{Benchmark}
We first run COLMAP with Doppelgangers++ to obtain pseudo ground-truth camera poses and scene structure. For each scene, we identify an anchor pair with low normal-pair probability and negligible COLMAP covisibility, which defines two visually similar but geometrically separated sides of the scene. Starting from one anchor, we construct a 15-view reference set using farthest-point sampling over COLMAP camera centers. Reference views are required to be consistent with the reference side and geometrically connected to the selected set. We then sample 5 Doppelganger views from the opposite side, choosing views that are visually similar to the reference set but geometrically inconsistent with it. To reduce redundancy, we remove duplicate or highly overlapping subsets and manually verify the retained samples. This procedure produces 43 subsets, each containing 15 reference views and 5 Doppelganger views.
\subsubsection{Camera Visualizations}
\begin{figure}
    \centering
    \includegraphics[width=0.8\linewidth]{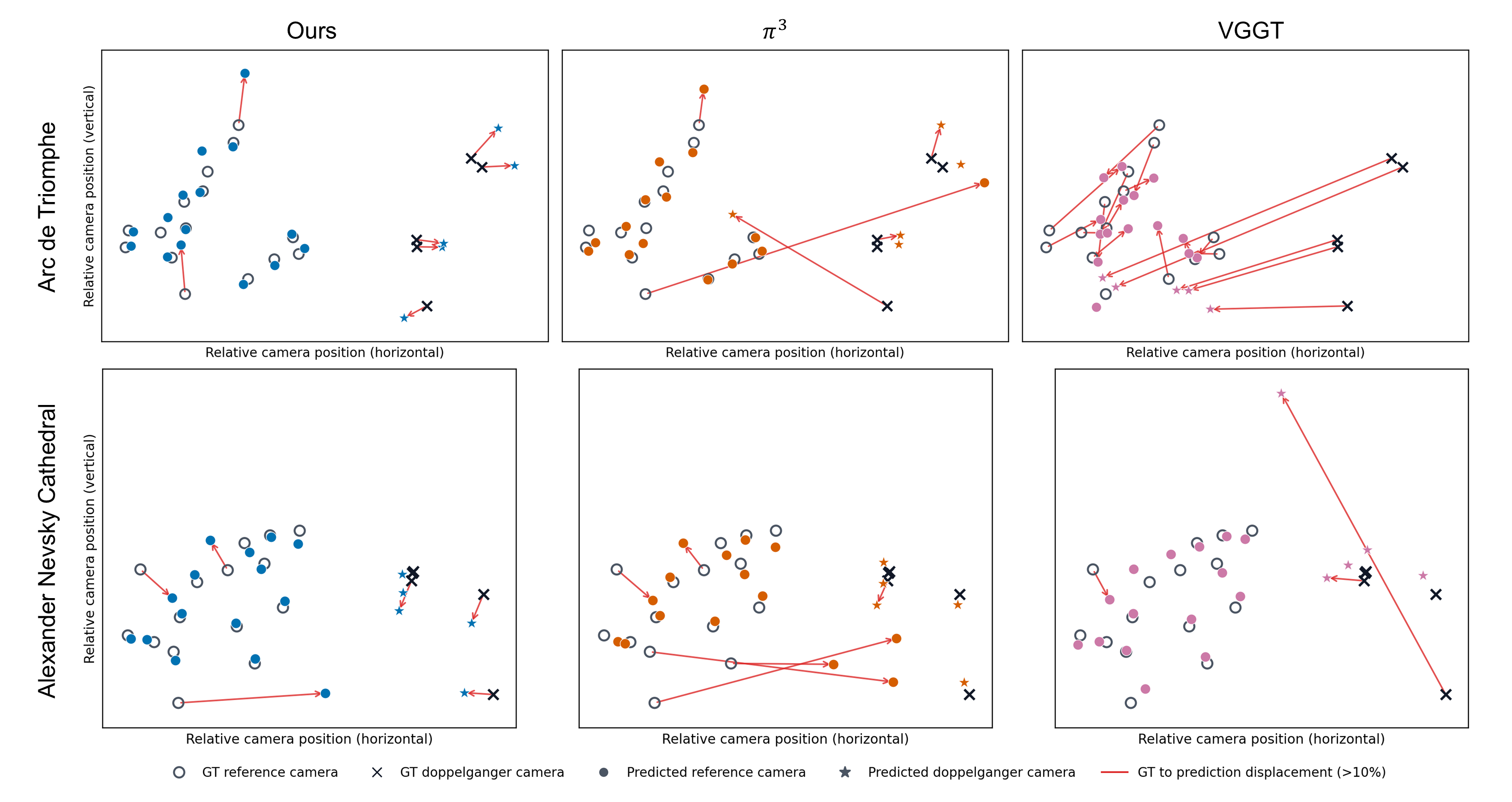}
    \caption{Visualizations of camera poses on visual ambiguity evaluation.}
    \label{fig:cameras}
\end{figure}
We visualize the predicted camera poses from two scenes by projecting them onto the 2D ground plane, as shown in Figure~\ref{fig:cameras}. VGGT and $\pi^3$ often place the reference and Doppelganger cameras on the same side of the scene whereas MGE better preserves the separation between the two geometrically distinct groups, consistent with its higher cross-pair pose accuracy.

\input{tables/appendix/longsequence}
\subsection{Quantitative metrics for token merging comparison}
\label{app:long_sequence}

We provide the detailed metrics in Table~\ref{tab:nrgbd_long_sequence} which are visualized in Figure~\ref{fig:long_seq}. Our method consistently outperforms existing long-sequence baselines across different keyframe strides. The advantage is especially pronounced under sparse keyframe sampling, where our model achieves substantially lower reconstruction errors and higher normal consistency than TTT3R, StreamVGGT, and FastVGGT. As the sequence becomes denser, our method remains robust and continues to deliver the best or comparable performance across all metrics. Notably, StreamVGGT runs out of memory for denser settings, whereas our method scales to all evaluated strides. We have a favorable balance between reconstruction quality and long-sequence scalability.

\input{tables/pointmap}
\input{tables/videodepth}

\subsection{Evaluation on More Standard benchmarks}
\label{app:conv}
\subsubsection{Point Map Estimation}
We evaluate the quality of reconstructed multi-view point maps on the scene-level 7-Scenes~\citep{shotton2013scene} and NRGBD~\citep{azinovic2022neural} datasets under both sparse and dense view conditions~\citep{wang2025continuous, wang2026pi}. 
Consistent with prior works~\citep{azinovic2022neural,Wang_2024_CVPR,wang2025continuous}, we report Accuracy (Acc.), Completion (Comp.), and Normal Consistency (NC) in Table~\ref{tab:mv_recon_cut3r}. Our MGE models achieve significant improvements on 7-scenes with the masked training especially under sparse input, and achieve comparative results on NRGBD to full-attention models.

\subsubsection{Video Depth Evaluation}
For the task of video depth estimation, we use the Sintel~\citep{NEURIPS2021_0a87257e}, Bonn~\citep{palazzolo2019refusion}), and KITTI~\citep{geiger2013vision} datasets, and report the Absolute Relative Error
(Abs Rel) and the prediction accuracy at a threshold of $\delta < $  1.25, following previous works~\citep{wang2025continuous, wang2026pi}. We align the video depth to the ground truth with a scale per sequence. As reported in Table~\ref{tab:videodepth}, compared with the full-attention baseline, our masked-trained model yields consistent improvements on Sintel and Bonn, achieving new SoTA on them while remaining competitive on KITTI.

\subsection{Ablation on Masking pattern}
\label{app:ablate_pattern}
\begin{figure}
\centering

\begin{minipage}[t]{0.45\linewidth}
    \vspace{0pt}
    \centering
    \includegraphics[width=\linewidth]{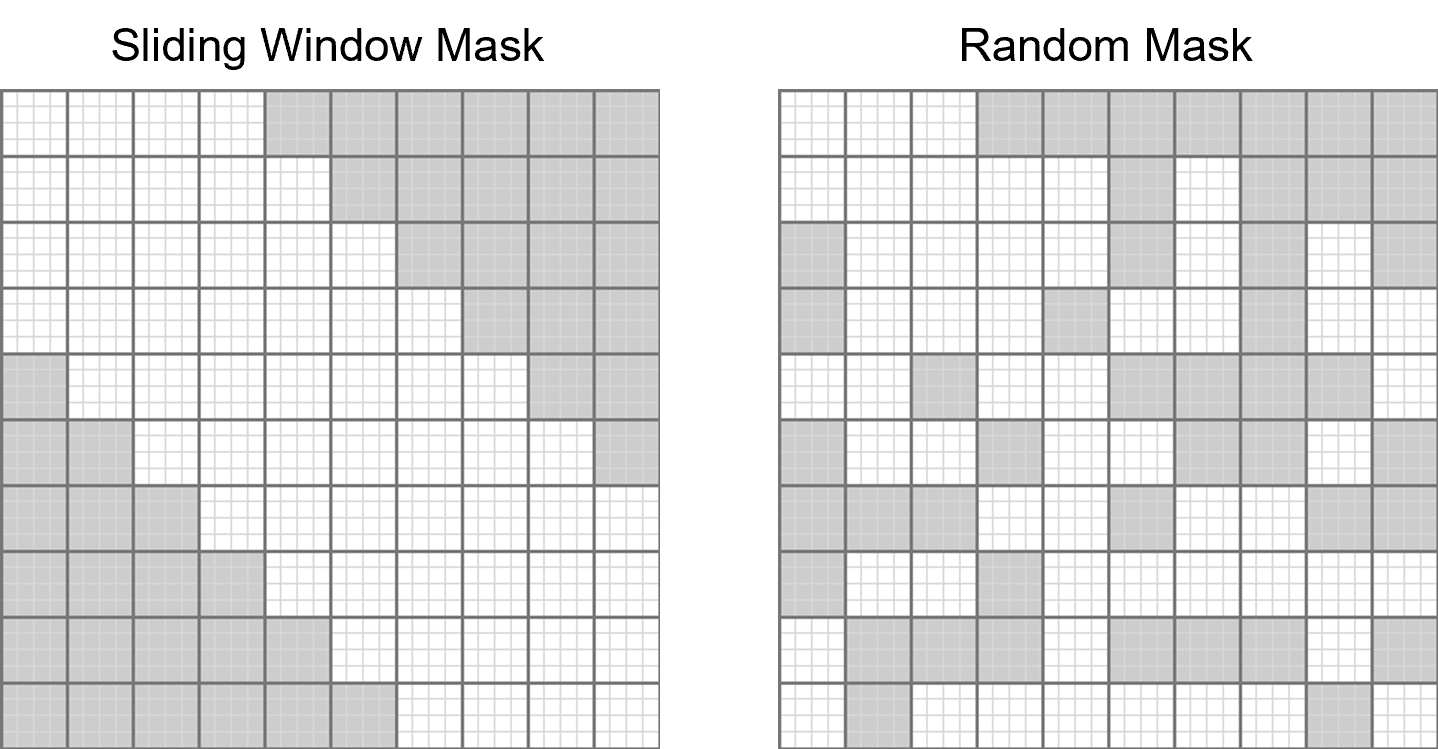}
    \captionof{figure}{Visualization of two mask patterns.}
    \label{fig:swa_mask}
\end{minipage}
\hfill
\begin{minipage}[t]{0.51\linewidth}
    \vspace{0pt}
    \centering
    \captionof{table}{\textbf{Ablation of mask pattern.} We compare the sliding-window attention (SWA) mask with the proposed random masking strategy under different keyframe strides.}
    \vspace{-0.5em}
    \resizebox{\linewidth}{!}{
    \begin{tabular}{lcccc}
        \toprule
        \multirow{2}{*}{Shape}
        & \multicolumn{2}{c}{\textbf{stride=100}}
        & \multicolumn{2}{c}{\textbf{stride=10}} \\
        \cmidrule(lr){2-3} \cmidrule(lr){4-5}
        & CD $\downarrow$ & NC $\uparrow$
        & CD $\downarrow$ & NC $\uparrow$ \\
        \midrule
        SWA
        & 0.0152 & 0.9125
        & 0.0135 & 0.7902 \\

        Random
        & \textbf{0.0145} & \textbf{0.9194}
        & \textbf{0.0117} & \textbf{0.8000} \\
        \bottomrule
    \end{tabular}
    }
    \label{tab:ablation_mask_pattern}
\end{minipage}
\end{figure}

We further compare random masking with Sliding-Window Attention (SWA) during masked training. The corresponding mask patterns are visualized in Figure~\ref{fig:swa_mask}, with comparable masking ratios used for a fair comparison. We evaluate both strategies on NRGBD~\citep{azinovic2022neural} using keyframe strides of 100 and 10, representing limited- and redundant-view settings, respectively. As shown in Table~\ref{tab:ablation_mask_pattern}, random masking consistently outperforms SWA under both settings, with a larger advantage for denser sequences. We attribute this to the limited propagation range of local sliding windows: similar to sequential matching, information must be relayed through intermediate views, whereas random connectivity enables more diverse long-range interactions and better approximates exhaustive multi-view reasoning.
\begin{figure}
    \centering
    \includegraphics[width=1\linewidth]{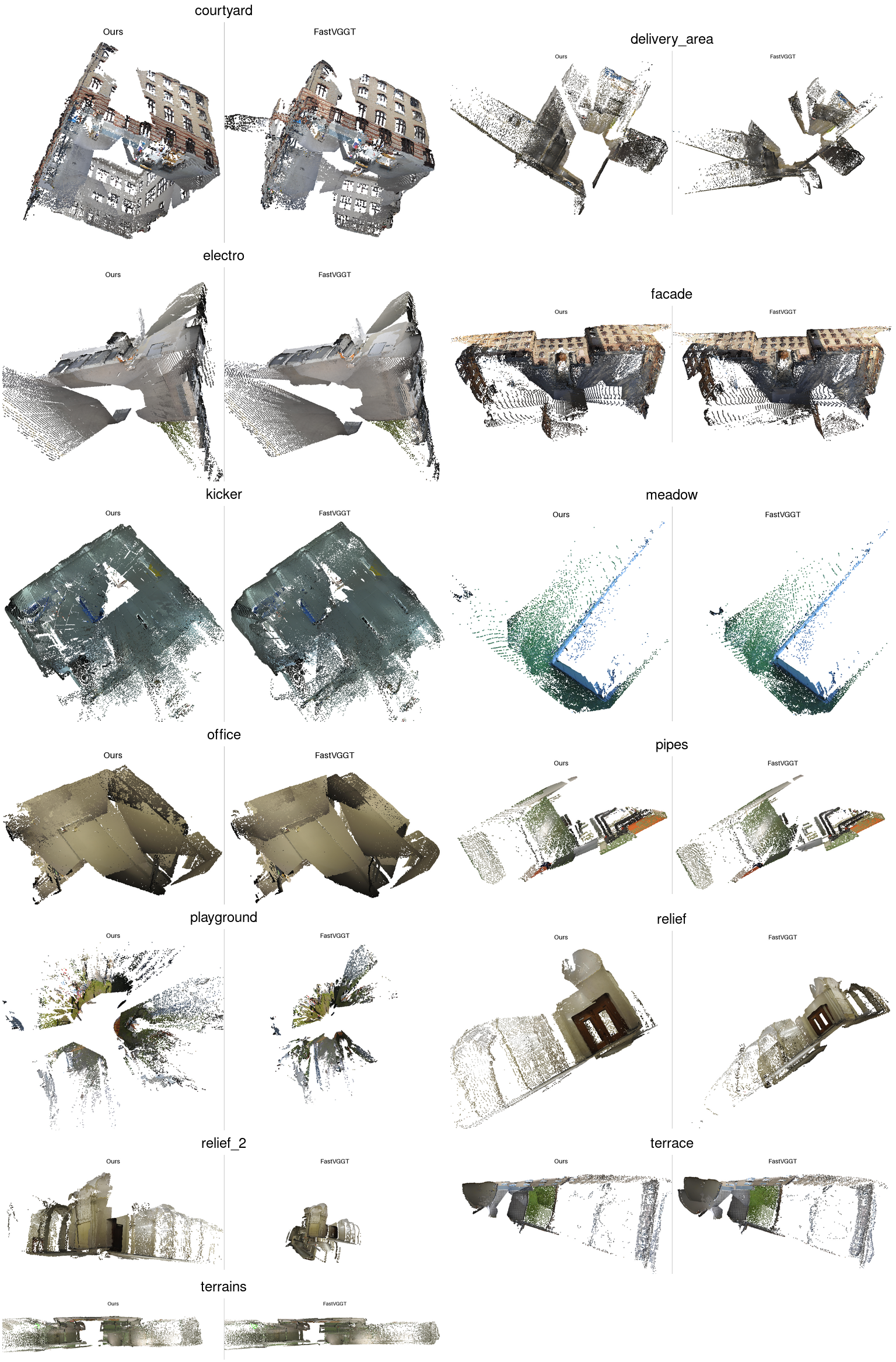}
    \caption{Qualitative comparison on ETH3D. For each scene, our reconstruction is shown on the left and FastVGGT on the right. Our method generally produces more complete and globally coherent geometry, with fewer missing surfaces and disconnected structures.}
    \label{fig:eth3d_qualitative}
\end{figure}

\subsection{Visualizations}
Figure~\ref{fig:eth3d_qualitative} presents qualitative reconstruction results on the 13 ETH3D scenes of our MGE+AGA and FastVGGT. Compared with FastVGGT, our method generally recovers more complete scene geometry while maintaining a coherent global structure. The improvement is particularly visible in challenging scenes such as courtyard, delivery area, and relief 2, where FastVGGT produces missing regions or disconnected fragments, whereas our reconstruction preserves a larger portion of the observed surfaces.

%% file: tables/appendix/longsequence.tex
\begin{table*}[t]
\centering
\caption{Dense reconstruction results under different keyframe strides. Lower is better for Acc, Comp, and CD, while higher is better for NC.}
\vspace{-5pt}
\resizebox{\linewidth}{!}{
\begin{tabular}{lcccccccccccc}
\toprule
\multirow{2}{*}{Method}
& \multicolumn{4}{c}{\textbf{stride=200}}
& \multicolumn{4}{c}{\textbf{stride=100}}
& \multicolumn{4}{c}{\textbf{stride=40}} \\
\cmidrule(lr){2-5}
\cmidrule(lr){6-9}
\cmidrule(lr){10-13}
& Acc $\downarrow$ & Comp $\downarrow$ & CD $\downarrow$ & NC $\uparrow$
& Acc $\downarrow$ & Comp $\downarrow$ & CD $\downarrow$ & NC $\uparrow$
& Acc $\downarrow$ & Comp $\downarrow$ & CD $\downarrow$ & NC $\uparrow$ \\
\midrule
TTT3R
& 0.097 & 0.064 & 0.0805 & 0.812
& 0.085 & 0.051 & 0.0680 & 0.811
& 0.079 & 0.033 & 0.0560 & 0.792 \\

StreamVGGT
& 0.092 & 0.083 & 0.0875 & 0.827
& 0.086 & 0.060 & 0.0730 & 0.820
& 0.090 & 0.052 & 0.0710 & 0.799 \\

FastVGGT
& 0.047 & 0.072 & 0.0595 & 0.869
& 0.034 & 0.047 & 0.0405 & 0.871
& 0.016 & 0.015 & 0.0155 & 0.843 \\

VGGT+AGA
& 0.041 & 0.054 & 0.0475 & 0.876
& 0.027 & 0.028 & 0.0275 & 0.879
& 0.016 & 0.014 & 0.0150 & 0.848 \\

\textbf{Ours}
& \textbf{0.021} & \textbf{0.022} & \textbf{0.0215} & \textbf{0.904}
& \textbf{0.018} & \textbf{0.018} & \textbf{0.0180} & \textbf{0.892}
& \textbf{0.014} & \textbf{0.014} & \textbf{0.0140} & \textbf{0.856} \\
\bottomrule
\end{tabular}
}
\vspace{2mm}

\resizebox{\linewidth}{!}{
\begin{tabular}{lcccccccccccc}
\toprule
\multirow{2}{*}{Method}
& \multicolumn{4}{c}{\textbf{stride=25}}
& \multicolumn{4}{c}{\textbf{stride=10}}
& \multicolumn{4}{c}{\textbf{stride=3}} \\
\cmidrule(lr){2-5}
\cmidrule(lr){6-9}
\cmidrule(lr){10-13}
& Acc $\downarrow$ & Comp $\downarrow$ & CD $\downarrow$ & NC $\uparrow$
& Acc $\downarrow$ & Comp $\downarrow$ & CD $\downarrow$ & NC $\uparrow$
& Acc $\downarrow$ & Comp $\downarrow$ & CD $\downarrow$ & NC $\uparrow$ \\
\midrule
TTT3R
& 0.081 & 0.034 & 0.0575 & 0.767
& 0.085 & 0.029 & 0.0570 & 0.706
& 0.123 & 0.042 & 0.0825 & 0.623 \\

StreamVGGT
& 0.096 & 0.052 & 0.0740 & 0.782
& \multicolumn{4}{c}{OOM}
& \multicolumn{4}{c}{OOM} \\

FastVGGT
& 0.019 & 0.017 & 0.0180 & 0.811
& 0.015 & 0.011 & 0.0130 & 0.734
& 0.016 & 0.010 & 0.0130 & 0.636 \\

VGGT+AGA
& 0.018 & 0.016 & 0.0170 & 0.811
& 0.014 & 0.012 & 0.0130 & 0.734
& 0.015 & 0.009 & 0.0120 & 0.637 \\

\textbf{Ours}
& \textbf{0.017} & \textbf{0.016} & \textbf{0.0165} & \textbf{0.829}
& \textbf{0.014} & \textbf{0.011} & \textbf{0.0125} & \textbf{0.763}
& \textbf{0.012} & \textbf{0.009} & \textbf{0.0105} & \textbf{0.678} \\
\bottomrule
\end{tabular}
}

\label{tab:nrgbd_long_sequence}
\end{table*}

%% file: tables/pointmap.tex
\begin{table}[htp]
    \centering
    \caption{
        Point map estimation on 7-Scenes and NRGBD.
    }
    \vspace{-1em}
    \resizebox{1.0\columnwidth}!{
    \begin{tabular}{lcccccccccccccc}
        \toprule[0.17em]
        {\multirow{4}{*}{\textbf{Method}}} &
        {\multirow{4}{*}{\textbf{View}}} &
        \multicolumn{6}{c}{\textbf{7-Scenes}} &
        \multicolumn{6}{c}{\textbf{NRGBD}} \\
        \cmidrule(r){3-8} \cmidrule(r){9-14}
        & &
        \multicolumn{2}{c}{Acc. $\downarrow$}  &
        \multicolumn{2}{c}{Comp. $\downarrow$} &
        \multicolumn{2}{c}{NC $\uparrow$}     & 
        \multicolumn{2}{c}{Acc. $\downarrow$}  &
        \multicolumn{2}{c}{Comp. $\downarrow$} &
        \multicolumn{2}{c}{NC $\uparrow$}     & \\
        \cmidrule(r){3-4} \cmidrule(r){5-6} \cmidrule(r){7-8}
        \cmidrule(r){9-10} \cmidrule(r){11-12} \cmidrule(r){13-14}
        & &
        Mean & Med. &
        Mean & Med. &
        Mean & Med. &
        Mean & Med. &
        Mean & Med. &
        Mean & Med. & \\
        \midrule[0.08em]
        Fast3R~\citep{yang2025fast3r} & \multirow{6}{*}{\textit{sparse}} & 0.095 & 0.065 & 0.144 & 0.089 & 0.673 & 0.759 & 0.135 & 0.091 & 0.163 & 0.104 & 0.759 & 0.877 \\
        CUT3R~\citep{wang2025continuous} & & 0.093 & 0.049 & 0.102 & 0.051 & 0.704 & 0.805 & 0.104 & 0.041 & 0.079 & 0.031 & 0.822 & 0.968 \\
        FLARE~\citep{zhang2025flare} & & 0.085 & 0.057 & 0.145 & 0.107 & 0.696 & 0.780 & 0.053 & 0.024 & 0.051 & 0.025 & 0.877 & 0.988 \\
        VGGT~\citep{wang2025vggt} & & 0.044 & \underline{0.025} & \underline{0.056} & \underline{0.033} & 0.733 & \underline{0.845} & 0.051 & 0.029 & 0.066 & 0.038 & 0.890 & 0.981 \\
       
        $\pi^3$~\citep{wang2026pi} & & 0.047 & 0.029 & 0.075 & 0.049 & \underline{0.742} & 0.841 & \textbf{0.026} & \textbf{0.015} & \textbf{0.028} & \textbf{0.014} & \textbf{0.916} & \textbf{0.992} \\
        \textbf{MGE-VGGT} & &\textbf{0.037} & \textbf{0.023} & \textbf{0.054} & \textbf{0.032} & 0743 & 0.851 & 0.050 & 0.028 & 0.060 & 0.033 & 0.891 & 0.990 \\ 
            \textbf{MGE-}$\bm{\pi^3}$ & & \underline{0.044} & 0.027 & 0.066 & 0.042 & \textbf{0.744} & \textbf{0.846} & \underline{0.027} & \underline{0.016} & \underline{0.030} & \underline{0.016} & \underline{0.901} & \underline{0.989} \\
        
        \midrule[0.08em]
        Fast3R~\citep{yang2025fast3r} & \multirow{6}{*}{\textit{dense}} & 0.040 & 0.017 & 0.056 & 0.018 & 0.644 & 0.725 & 0.072 & 0.030 & 0.050 & 0.016 & 0.790 & 0.934 \\
        CUT3R~\citep{wang2025continuous} & & 0.023 & 0.010 & 0.027 & \textbf{0.008} & 0.669 & 0.764 & 0.086 & 0.037 & 0.048 & 0.017 & 0.800 & 0.953 \\
        FLARE~\citep{zhang2025flare} & & 0.019 & \underline{0.007} & 0.026 & 0.013 & 0.684 & 0.785 & 0.023 & 0.011 & 0.018 & 0.008 & 0.882 & 0.986 \\
        VGGT~\citep{wang2025vggt} & & 0.022 & 0.008 & 0.026 & 0.012 & 0.666 & 0.760 & 0.017 & 0.010 & 0.015 & \underline{0.005} & 0.893 & \textbf{0.988} \\
        
        $\pi^3$~\citep{wang2026pi} & & \underline{0.016} & \underline{0.007} & \textbf{0.022} & 0.011 & \textbf{0.689} & \textbf{0.792} & \underline{0.015} & \underline{0.008} & \textbf{0.013} & \textbf{0.005} & \textbf{0.898} & 0.987 \\
        \textbf{MGE-VGGT} & & 0.019 & 0.008 & 0.027 & 0.009 & 0.683 & 0.784  & 0.014 & 0.007 &0.016 & 0.006 & 0.889 & 0.987 \\
        \textbf{MGE-}$\bm{\pi^3}$ & & \textbf{0.016} & \textbf{0.007} & \underline{0.022} & \underline{0.010} & \underline{0.687} & \underline{0.789} & \textbf{0.014} & \textbf{0.007} & \underline{0.014} & 0.006 & \underline{0.897} & \underline{0.988} \\
        
        \bottomrule[0.17em] 
    \end{tabular}
    }
    \label{tab:mv_recon_cut3r}
\end{table}

%% file: tables/videodepth.tex
\begin{table}
    \centering
    \vspace{-1em}
    \caption{
        Video depth estimation on Sintel, Bonn and KITTI.
    }
    \vspace{-1em}
    \resizebox{1.0\columnwidth}!{
    \begin{tabular}{lccccccc}
        \toprule[0.17em]
        {\multirow{3}{*}{\textbf{Method}}} &
        \multicolumn{2}{c}{\textbf{Sintel}} &
        \multicolumn{2}{c}{\textbf{Bonn}} &
        \multicolumn{2}{c}{\textbf{KITTI}}\\
        \cmidrule(r){2-3} \cmidrule(r){4-5} \cmidrule(r){6-7}
        &
        Abs Rel $\downarrow$ & $\delta<1.25$ $\uparrow$ &
        Abs Rel $\downarrow$ & $\delta<1.25$ $\uparrow$ &
        Abs Rel $\downarrow$ & $\delta<1.25$ $\uparrow$ \\
        \midrule[0.08em]
        DUSt3R~\citep{Wang_2024_CVPR}      & 0.662 & 0.434 & 0.151 & 0.839 & 0.143 & 0.814 \\
        MASt3R~\citep{leroy2024grounding}  & 0.558 & 0.487 & 0.188 & 0.765 & 0.115 & 0.848 \\
        MonST3R~\citep{zhang2025monst3r}   & 0.399 & 0.519 & 0.072 & 0.957 & 0.107 & 0.884\\
        Fast3R~\citep{yang2025fast3r}      & 0.638 & 0.422 & 0.194 & 0.772 & 0.138 & 0.834 \\
        MVDUSt3R~\citep{tang2025mv}        & 0.805 & 0.283 & 0.426 & 0.357 & 0.456 & 0.342 \\
        CUT3R~\citep{wang2025continuous}  & 0.417 & 0.507 & 0.078 & 0.937 & 0.122 & 0.876 \\
        Aether~\citep{zhu2025aether}           & 0.324 & 0.502 & 0.273 & 0.594 & 0.056 & 0.978 \\
        FLARE~\citep{zhang2025flare}    & 0.729 & 0.336 & 0.152 & 0.790 & 0.356 & 0.570 \\
        VGGT~\citep{wang2025vggt}    & 0.299 & 0.638 & 0.057 & 0.966 & 0.062 & 0.969 \\
        
        $\pi^3$~\citep{wang2026pi}        &  \underline{0.233} &  \underline{0.664} & \underline{0.049} & \underline{0.975} & \textbf{0.038} & \textbf{0.986}\\
        \midrule[0.08em]
        \textbf{MGE-VGGT} & 0.283 & 0.663 & 0.054 & 0.970 & 0.061 & 0.972 \\
        \textbf{MGE-}$\bm{\pi^3}$        & \textbf{0.232} & \textbf{0.676} & \textbf{0.049} & \textbf{0.975} & \underline{0.038} & \underline{0.985}\\
        \bottomrule[0.17em]
    \end{tabular}
    }
    \label{tab:videodepth}
\end{table}